\documentclass[]{bytedance_seed}

\makeatletter 
\AtBeginDocument{%
  \setlength{\headheight}{34pt}%
  \setlength{\headsep}{18pt}%
}

\let\old@maketitle\maketitle
\renewcommand{\maketitle}{%
  \old@maketitle
  \vspace{8pt} 
}
\makeatother

\usepackage[T1]{fontenc}
\usepackage[utf8]{inputenc}

\usepackage{amsmath}
\usepackage{amsfonts}
\usepackage{amssymb}
\usepackage{bbm}

\usepackage{graphicx}
\usepackage{subcaption}
\usepackage{wrapfig}
\graphicspath{{figure/}}

\usepackage{booktabs}
\usepackage{multirow}
\usepackage{threeparttable}
\usepackage{tabularx}
\usepackage{array}
\usepackage[table]{xcolor}
\usepackage{wrapfig}
\usepackage[ruled]{algorithm2e}
\usepackage{algpseudocode}

\usepackage{microtype}
\usepackage{float}
\usepackage{enumitem}
\usepackage{multicol}
\usepackage[table]{xcolor}   
\definecolor{bestblue}{RGB}{222,236,255} 
\newcommand{\bestcell}[1]{\cellcolor{bestblue}\textbf{#1}} 
\usepackage{makecell}

\usepackage{hyperref}
\usepackage{url}

\definecolor{osured}{RGB}{187,0,0}

\definecolor{SeedBlue}{RGB}{187,0,0}
\definecolor{PrimaryColor}{RGB}{187,0,0}
\definecolor{AccentColor}{RGB}{187,0,0}
\colorlet{structurecolor}{osured}

\usepackage{tcolorbox}

\newtcolorbox{promptbox}[2][Judge Prompt]{
  colback=black!5!white,
  arc=5pt,
  boxrule=0.6pt,
  fonttitle=\bfseries,
  title=#1,
  before upper={\small},
  colframe=osured,
  label=#2,
}



\title{MMDeepResearch-Bench: \\ A Benchmark for  Multimodal Deep Research Agents}

\author{
Peizhou Huang*,
Zixuan Zhong*,
Zhongwei Wan*,
Donghao Zhou*, 
Samiul Alam, 
Xin Wang,
Zexin Li, 
Zhihao Dou, 
Li Zhu, 
Jing Xiong,
Chaofan Tao, 
Yan Xu, 
Dimitrios Dimitriadis,
Tuo Zhang, Mi Zhang
}

\affiliation{%
  \textsuperscript{*} Equal Contribution\\
  OSU, Amazon, UMich, UCL, CUHK, UCR, CWRU, HKU\\[3pt]
  {\textbf{Correspondence:} Tuo Zhang \href{mailto:tuozhang@amazon.com}{\texttt{tuozhang@amazon.com}},
  Mi Zhang \href{mailto:mizhang.1@osu.edu}{\texttt{mizhang.1@osu.edu}}}\\
  {\textbf{Project Page:} \url{https://mmdeepresearch-bench.github.io}}%
}

\abstract{
Deep Research Agents (DRAs) generate citation-rich reports via multi-step search and synthesis,
yet assessing faithful use of textual and visual evidence in open-ended reports remains challenging.
We introduce \textbf{MMDeepResearch-Bench (MMDR-Bench)},
a benchmark of 140 researcher-curated tasks across 19 domains,
where each task provides an image--text bundle to evaluate multimodal understanding
and citation-grounded report generation.
Compared to prior setups, MMDR-Bench emphasizes report-style synthesis with explicit evidence use,
where models must connect visual artifacts to sourced claims
and maintain consistency across narrative, citations, and visual references.
We further propose a unified, interpretable evaluation pipeline:
\textbf{F}ormula--LLM \textbf{A}daptive \textbf{E}valuation (\textbf{FLAE}) for report quality,
\textbf{T}rustworthy \textbf{R}etrieval-\textbf{A}ligned \textbf{C}itation \textbf{E}valuation (\textbf{TRACE}) for citation-grounded evidence alignment,
and \textbf{M}ultimodal \textbf{S}upport-\textbf{A}ligned \textbf{I}ntegrity \textbf{C}heck (\textbf{MOSAIC}) for text--visual integrity,
each producing fine-grained signals for aggregate failure-mode analysis beyond a single overall score.
Experiments across 28 state-of-the-art models and agents reveal systematic trade-offs between
generation quality, citation discipline, and multimodal grounding,
highlighting that strong prose alone does not guarantee faithful evidence use
and that multimodal integrity remains a key bottleneck for deep research agents.
\href{https://huggingface.co/datasets/MMDR-2025/MMdeepresearch}{Dataset};
\href{https://github.com/AIoT-MLSys-Lab/MMDeepResearch-Bench}{evaluation code}.
}

\begin{document}
\maketitle
\suppressfloats[t]

\section{Introduction}

Recent advancements in foundation models have driven a shift from language-centric systems to large multimodal models (LMMs) that jointly process text and visual inputs \citep{bommasani2021foundation}. Enabled by vision--language pretraining and instruction tuning \citep{blip,instructblip,minigpt}, modern LMMs can reason over structured visual artifacts such as charts and documents, forming the basis of current vision--language benchmarks \citep{Qwen3-VL,chartqa,docvqa,chartmuseum}.

Yet static models remain limited by fixed parametric memory, motivating retrieval-augmented generation \citep{RAG} and tool-using agents that browse and collect external evidence \citep{wong2025widesearch,xi2025infodeepseek,lan2025deepwidesearch}. Building on this paradigm, Deep Research Agents (DRAs) target open-ended, long-horizon tasks by iteratively retrieving sources, reconciling hypotheses, and producing research-style reports \citep{DPsurvey,storm,du2025deepresearch}. Since real research is rarely text-only, DRAs must also align textual claims with figures, charts, and diagrams, motivating multimodal deep research \citep{du2025deepresearch,patel2025deepscholar,venkit2025deeptrace}.

As DRAs proliferate \citep{patel2025deepscholar,venkit2025deeptrace}, evaluation becomes crucial but difficult: intermediate reasoning and retrieval are opaque, and open-ended questions seldom admit a single gold answer, making the final cited report the primary evaluation interface \citep{venkit2025deeptrace}. Existing benchmarks either isolate web or retrieval competence \citep{wong2025widesearch,xi2025infodeepseek}, focus on text-only deep research reports \citep{du2025deepresearch,patel2025deepscholar}, or emphasize short-horizon multimodal perception \citep{zou2025uni,chartmuseum}. Live retrieval further complicates evaluation via issues such as search-time data contamination \citep{han2025search}. Recent benchmarks cover complementary targets: visual--textual search and answer correctness in VDR-Bench \citep{zeng2026vdr}, expert-derived report rubrics in DeepResearch Bench II \citep{li2026drbii}, and synthesis, factuality, and process audits in MiroEval \citep{ye2026miroeval}. MMDR-Bench jointly evaluates report quality, citation support, input-image fidelity, and report-image alignment.

\begin{figure}[t]
  \centering
  \includegraphics[width=0.78\linewidth]{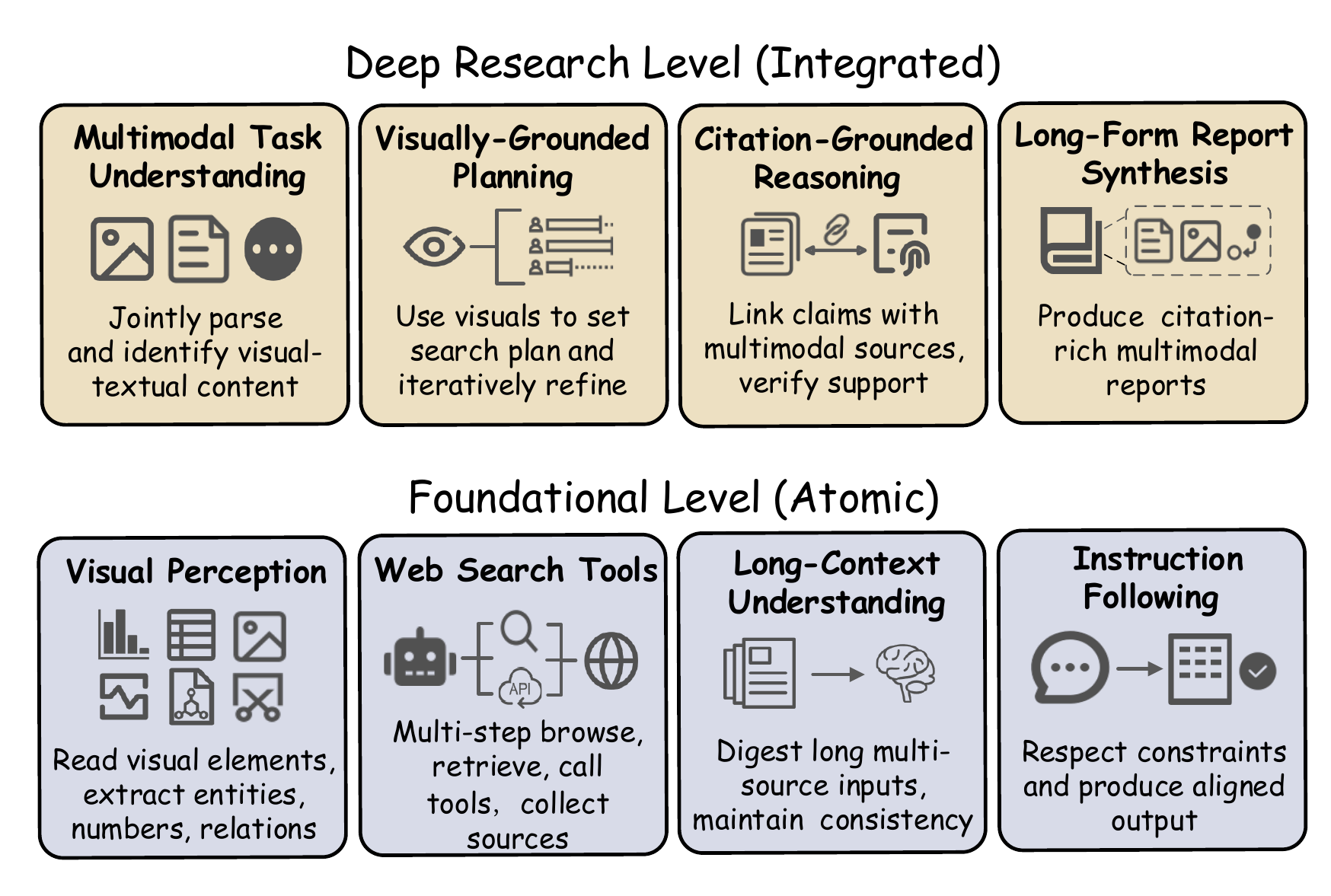}
  \vspace{-3mm}
  \caption{An overview of MMDR-Bench.}
  \label{fig:first}
  \vspace{-3mm}
\end{figure}

\begin{figure}[t]
  \centering
  \includegraphics[width=0.98\linewidth]{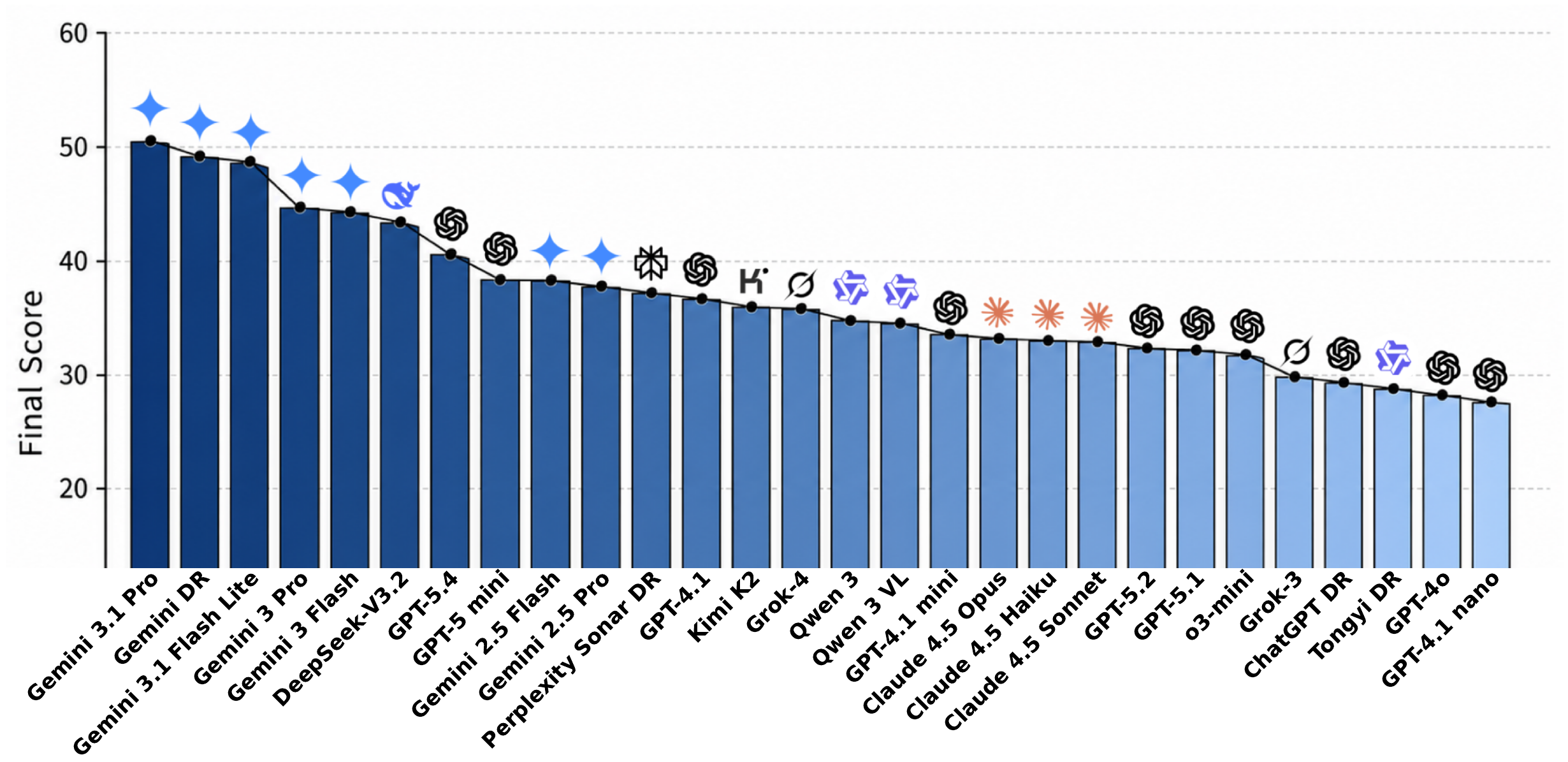}
  \caption{Overall scores for 28 systems on MMDR-Bench. Gemini DR denotes Gemini Deep Research (Gemini 3 Pro).}
\label{fig:intro-dim-and-results}
 \vspace{-3mm}
\end{figure}

We introduce \textbf{MMDeepResearch-Bench (MMDR-Bench)}, comprising 140 image--text tasks that elicit citation-grounded reports, integrating multimodal understanding, evidence-grounded reasoning, and long-form synthesis (Figure~\ref{fig:first}). MMDR-Bench includes two complementary regimes, \textit{Daily} and \textit{Research}, reflecting lightweight everyday usage and analysis-heavy research settings. All tasks are iteratively refined by researchers with relevant disciplinary training to ensure multimodal necessity and verifiability.

Building on MMDR-Bench, we develop an evaluation framework along three aspects: \textbf{F}ormula--\textbf{L}LM \textbf{A}daptive \textbf{E}valuation (\textbf{FLAE}) for long-form report quality, \textbf{T}rustworthy \textbf{R}etrieval--\textbf{A}ligned \textbf{C}itation \textbf{E}valuation (\textbf{TRACE}) for citation-grounded support and source quality, and \textbf{M}ultimodal \textbf{S}upport--\textbf{A}ligned \textbf{I}ntegrity \textbf{C}heck (\textbf{MOSAIC}) for text--image consistency of the report. Our contributions are summarized as follows.

\begin{itemize}[leftmargin=1.5em]
    \item \textbf{A Multimodal Research Benchmark.} MMDR-Bench contains 140 researcher-curated tasks across 19 domains under Daily and Research regimes, with image--text inputs requiring multimodal understanding and evidence-grounded report synthesis.
    \item \textbf{A Unified Evaluation Framework.} FLAE, TRACE, and MOSAIC jointly assess report quality, citation-grounded reasoning, and consistency between textual claims and visual artifacts.
    \item \textbf{Visual Evidence Fidelity.} VEF checks task-specific textualized visual requirements with a strict PASS/FAIL rule, penalizing misinterpretation of critical input-image evidence.
    \item \textbf{Systematic Evaluation and Open Resources.} Results across 28 models and agentic systems expose trade-offs between writing quality, citation discipline, and multimodal grounding. We release the dataset, evaluation code, and detailed metrics.
\end{itemize}

\FloatBarrier

\section{Related Work}

\noindent \textbf{Deep Search and Agentic Reasoning.}
Agentic search frameworks decompose queries and gather evidence through sequential tool use \citep{storm,deng2023mind2web}. Search-R1 \citep{jin2025search} and DeepDive \citep{lu2025deepdive} use reinforcement learning to improve retrieval and reasoning. These systems motivate evaluating not only final answers but also how reports substantiate their claims.

\noindent \textbf{Multimodal Search and Reasoning.}
MMSearch \citep{jiangmmsearch} and DeepEyes \citep{zheng2025deepeyes} study visual information seeking. Vision-DeepResearch Benchmark \citep{zeng2026vdr} targets realistic visual--textual search with 2,000 VQA instances and a cropped-search workflow. Our focus is instead long-form reports, including whether input-image facts and report images support the accompanying claims.
\par

\noindent \textbf{Benchmarks and Evaluation.}
BrowseComp \citep{wei2025browsecomp} and BrowseComp-Plus \citep{chen2025browsecomp} are text-based search benchmarks; the latter provides a fixed-corpus setting. DeepResearch Bench \citep{du2025deepresearch} and DeepScholar \citep{patel2025deepscholar} evaluate report synthesis. Concurrent DeepResearch Bench II \citep{li2026drbii} derives fine-grained rubrics from expert reports, while MiroEval \citep{ye2026miroeval} combines synthesis, factuality, and process-centric evaluation. MMDR-Bench complements these efforts with image--text task bundles and separate checks for input-image fidelity, citation support, and report-image alignment.

\section{Dataset Collection}

We define multimodal deep research as tasks that require multi-round web browsing, evidence gathering, and report synthesis while explicitly interpreting and using provided images. Following this definition, MMDR-Bench targets realistic long-horizon cases that require multimodal understanding and evidence-grounded report writing. Each task is an image--text bundle: a textual query paired with a small set of images that must be interpreted and integrated into a research report. This design jointly evaluates (i) multimodal question understanding and (ii) evidence-grounded multimodal report generation with citations.

\begin{figure}[t]
  \centering
  \begin{minipage}[b]{0.43\textwidth}
    \centering
    \includegraphics[width=\linewidth]{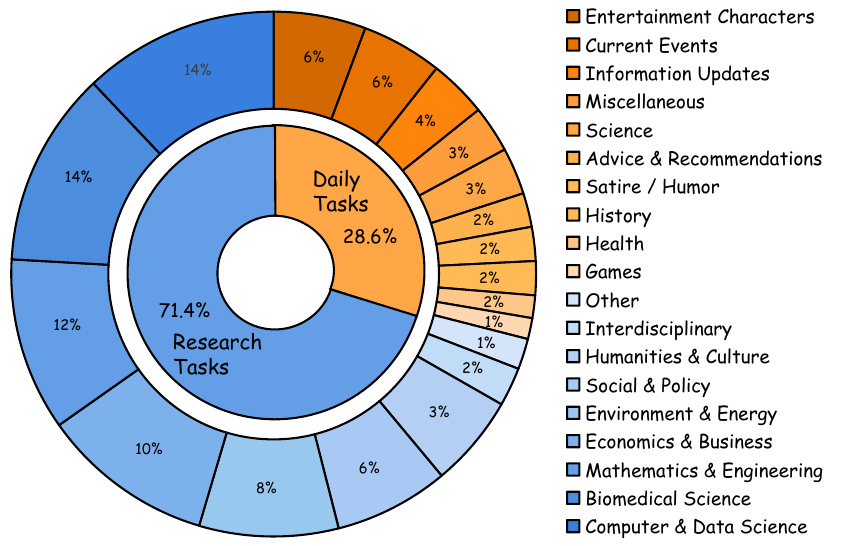}
    \caption{Task distribution of MMDR-Bench.}
    \label{fig:dataset-dist}
  \end{minipage}\hfill
  \begin{minipage}[b]{0.55\textwidth}
    \centering
    \includegraphics[width=\linewidth]{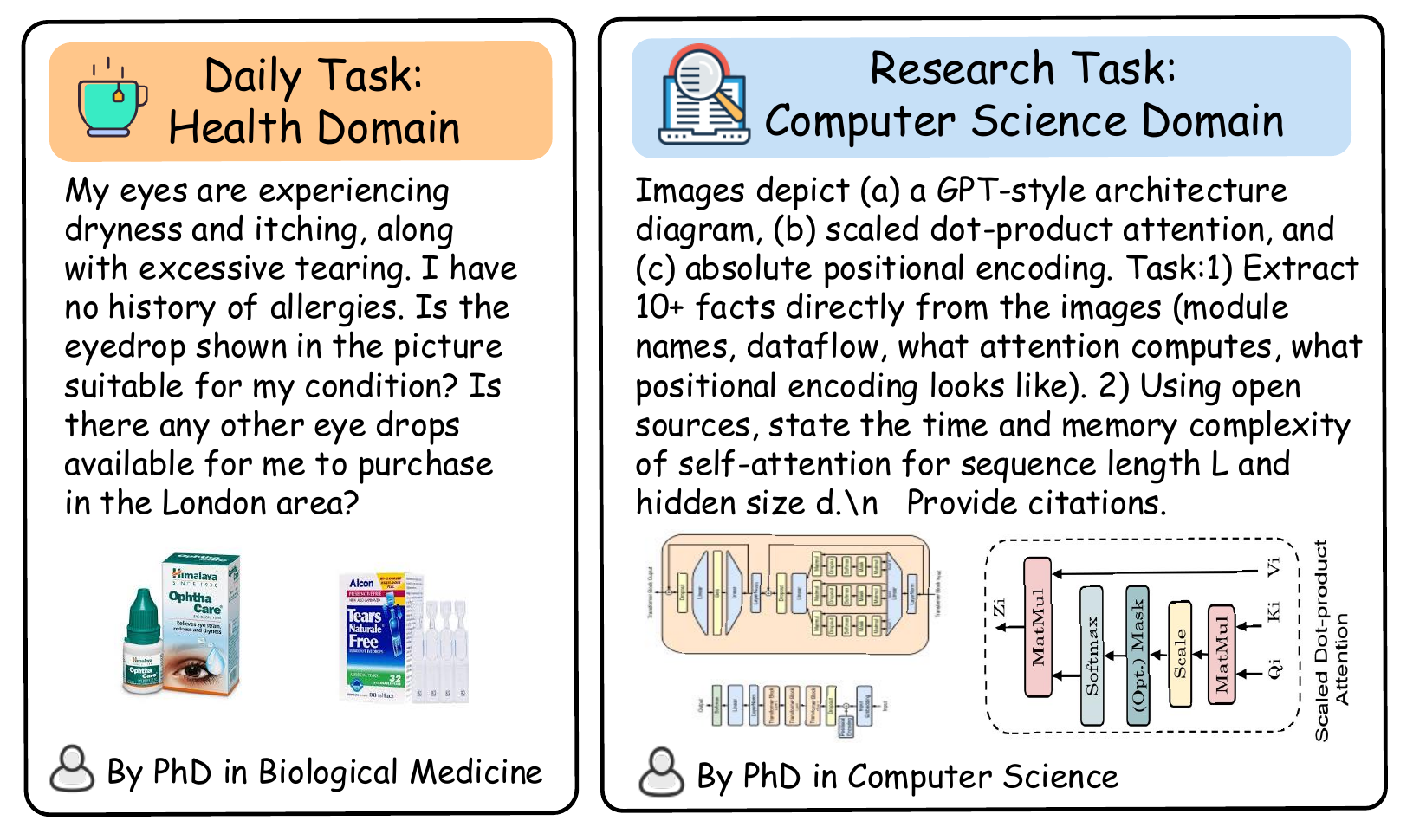}
    \caption{Two example tasks from MMDR-Bench.}
    \label{fig:dataset-examples}
  \end{minipage}
\end{figure}

From a collected pool of 98k+ real-world queries, author-team researchers with relevant disciplinary training curate 140 tasks across 19 domains (Figure~\ref{fig:dataset-dist}). The final tasks comprise two regimes: \textit{Daily} (40 tasks) with casual visuals such as screenshots and UI captures, and \textit{Research} (100 tasks) with structured, information-dense figures such as charts, tables, and diagrams that require deeper synthesis. Curation emphasizes multimodal necessity, answerability, public evidence, and auditability. Two example tasks are shown in Figure~\ref{fig:dataset-examples}; further construction details are in Appendix~\ref{app:dataset-details}.

\FloatBarrier

\section{Evaluation Methodology}
\label{sec:method}
\begin{figure}[!t]
  \centering
  \includegraphics[width=0.93\textwidth]{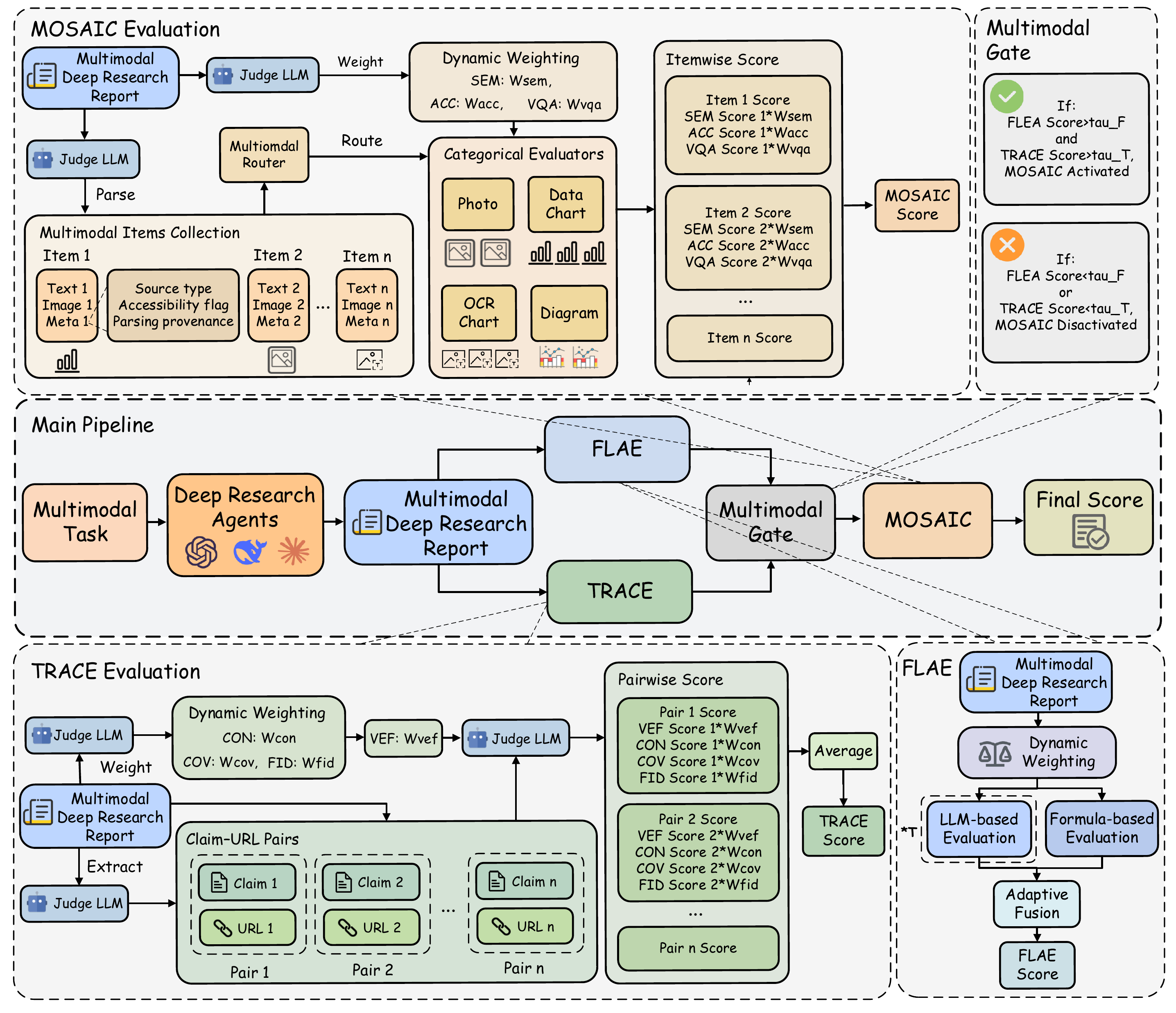}
  \caption{MMDR-Bench evaluation pipeline, integrating report-quality assessment, citation and visual-evidence verification, and gated visual-support evaluation.}
  \label{fig:Architecture}
\end{figure}

Our evaluation framework jointly assesses research-report quality, citation-grounded evidence use, and multimodal integrity. It comprises FLAE, TRACE, and MOSAIC (Figure~\ref{fig:Architecture}).

FLAE assesses report quality, TRACE verifies citation support and fidelity to task-provided visual evidence, and MOSAIC checks the images used in the generated report. Positive report-quality and text-evidence scores gate MOSAIC evaluation. Together, these modules connect writing quality to the evidence supporting a report's claims. Execution and aggregation details are provided in Appendix~\ref{app:reliability}.

\subsection{FLAE: Formula-LLM Adaptive Evaluation}
\label{sec:race-g}
Writing requirements vary across research tasks. \textbf{FLAE} (\textbf{F}ormula--\textbf{L}LM \textbf{A}daptive \textbf{E}valuation) combines fixed feature transforms with holistic judge scores on four dimensions: Readability (\textsc{Read.}), Insightfulness (\textsc{Insh.}), Coherence (\textsc{Coh.}), and Structural Completeness (\textsc{Stru.}).

\noindent \textbf{Formula-Based Channel.}
The feature vector $\phi(R)$ combines directly measured text statistics with judge-estimated attributes. Fixed transforms produce scores in $[0,1]$:
\begin{equation}
\label{eq:flae-form}
s^{\mathrm{form}}_d(R)=f_d\!\bigl(\phi(R)\bigr),\qquad
\mathcal D=\{\textsc{Read.},\textsc{Insh.},\textsc{Coh.},\textsc{Stru.}\}.
\end{equation}
The formula channel captures explicit report characteristics, complementing the judge's holistic assessment. Feature definitions and transforms are provided in Appendix~\ref{app:flae-formulas}.

\noindent \textbf{Judge Channel and Fusion.}
A judge estimates holistic scores $s^{\mathrm{judge}}_d(t,R)$. We fuse the channels as $s_d=\alpha\,s^{\mathrm{form}}_d+(1-\alpha)s^{\mathrm{judge}}_d$. The fusion coefficient $\alpha$ is set by report type, while task-adaptive dimension weights account for differences in writing requirements. This combination balances explicit features with holistic assessment; detailed settings are provided in Appendix~\ref{app:flae-fusion-prompts}.

\begin{equation}
\label{eq:flae-gen}
\mathrm{FLAE}(t,R)=100\sum_{d\in\mathcal D}W_d(t,R)s_d(R),\qquad \sum_d W_d=1.
\end{equation}

\subsection{TRACE: Trustworthy Retrieval-Aligned Citation Evaluation}
\label{sec:trace-evidence}
\textbf{TRACE} (\textbf{T}rustworthy \textbf{R}etrieval-\textbf{A}ligned \textbf{C}itation \textbf{E}valuation) assesses citation support and task-image fidelity. We map citation indices to URLs, extract atomic Claim--URL pairs, retrieve source content, and record accessibility. A judge assesses support, hallucination, causal inversion, and reasoning-chain soundness. Text-evidence aggregation uses four dimensions in $[0,1]$: Consistency (\textsc{Con.}), Coverage (\textsc{Cov.}), Textual Fidelity (\textsc{Fid.}), and Source Diversity (\textsc{Div.}).

Visual Evidence Fidelity (\textsc{Vef.}) checks whether the report faithfully uses the task-provided images. It compares the report against task-specific textualized visual ground truth and applies a strict binary criterion to critical visual facts, including entity identities, object presence, and numerical details. This makes fidelity to the input images an explicit part of evidence evaluation. Ground-truth construction and the decision rule are detailed in Appendix~\ref{app:vef-gt}.

With $\mathcal K=\{\textsc{Con.},\textsc{Cov.},\textsc{Fid.},\textsc{Div.}\}$, TRACE is
\begin{equation}
\label{eq:trace-score}
\mathrm{TRACE}(t,R)=100\left[\lambda_{\textsc{Vef.}}y(t,R)
 +(1-\lambda_{\textsc{Vef.}})\sum_{k\in\mathcal K}W_k(t,R)E_k(R)\right],
\end{equation}
where $y(t,R)$ is the binary VEF outcome, $\lambda_{\textsc{Vef.}}=0.4$, and $\sum_kW_k=1$. The citation weights use a fixed prior, task-controller adjustments, and document diagnostics (Appendix~\ref{app:trace}).

\subsection{MOSAIC: Multimodal Support-Aligned Integrity Check}
\label{sec:mosaic-mm}
\textbf{MOSAIC} (\textbf{M}ultim\textbf{o}dal \textbf{S}upport-\textbf{A}ligned \textbf{I}ntegrity \textbf{C}heck) evaluates whether image-referenced statements match the images embedded or linked in a report. It extracts multimodal items and routes them into photo, data-chart, OCR-chart, or diagram types for modality-appropriate checks.

\noindent \textbf{Multimodal Support Scoring.}
Applicable diagnostics include Visual-Semantic Alignment (\textsc{Sem.}), Visual Data Interpretation Accuracy (\textsc{Acc.}), and Visual Claim Support (\textsc{VQA}). VQA uses visual question answering to assess whether an image supports its associated statement. This targets the specific claim made in the report, complementing overall visual-semantic alignment. Item scores use fixed type-specific weights, a formatting/integration factor $f_i$, and a diagram hallucination factor $h_i$:
\begin{equation}
\label{eq:mosaic-item-score}
s_i=f_i\sum_k\omega_{\mathrm{type}(i),k}\,m_{i,k},\qquad \sum_k\omega_{\mathrm{type}(i),k}=1.
\end{equation}
For diagrams, $m_i$ includes $1-h_i$; other types use applicable \textsc{Sem.}, \textsc{Acc.}, and \textsc{VQA} scores. The report-level \textsc{MOSAIC} score is $100$ times the mean of $s_i$ over scored items and is set to $0$ when no item can be scored. Appendix~\ref{app:mosaic} gives the weights and missing-value conventions.

\subsection{Design Rationale: Artifact-Grounded Evaluation}
\label{sec:design-rationale}
MMDR-Bench evaluates three complementary aspects of multimodal deep research: faithful use of the provided visual inputs (\textsc{Vef.}), grounding of textual claims in retrieved evidence (\textsc{TRACE}), and consistency between a report's statements and its visual artifacts (\textsc{MOSAIC}). Open-ended research permits multiple valid evidence paths, while live sources and time-sensitive information continue to evolve. Rather than prescribing a single gold report or search route, our framework checks the evidence underlying the generated report. This accommodates diverse research strategies while making failures in visual interpretation, citation support, and multimodal synthesis visible through separate diagnostic dimensions.

\section{Experiments}
\label{sec:exp}

\subsection{Experimental Setup}
\label{sec:exp-setup}

\noindent \textbf{Benchmark and Protocol.} We evaluate generated reports on MMDR-Bench using Gemini-2.5-Pro as the primary judge. Overall scores combine \textsc{FLAE}, \textsc{TRACE}, and \textsc{MOSAIC} with weights of 20/50/30\%, emphasizing evidence-grounded research quality. Model access information and testing windows appear in Appendix~\ref{app:model_details}; judge settings, coverage, and missing-output handling are detailed in Appendix~\ref{app:reliability}.

\noindent \textbf{Evaluated Report Models.}
We evaluate a diverse set of systems spanning four groups: (i) single-modal LLM baselines without web search function, (ii) multimodal LLM baselines without web search function, (iii) multimodal LLM baselines with web search function and (iv) specific deep research agents. For each model, we report the overall MMDR-Bench score and metric dimensions for FLAE, TRACE and MOSAIC.

\subsection{Main Results and Findings} \label{sec:exp-main} 

\begin{table}[!t]
\caption{Results for 28 systems on MMDR-Bench, with Overall scores and selected diagnostic dimensions (0--100). Highlights indicate the highest scores in each column. MOSAIC submetrics average applicable task-level summaries, with system-dependent coverage (ACC: 6 tasks for o3-mini; 45 for Kimi K2). Aggregation and coverage details are provided in Appendix~\ref{app:reliability}.}
\label{main}
\label{tab:full-metric-breakdown}
\centering
\footnotesize
\setlength{\tabcolsep}{2.2pt}
\renewcommand{\arraystretch}{1.03}

\resizebox{\textwidth}{!}{%
\begin{tabular}{@{}l c ccc cccc ccc@{}}
\toprule
\multirow{2}{*}{Model} &
\multirow{2}{*}{Overall} &
\multicolumn{3}{c}{\textsc{FLAE}} &
\multicolumn{4}{c}{\textsc{TRACE}} &
\multicolumn{3}{c}{\textsc{MOSAIC}} \\
\cmidrule(lr){3-5}\cmidrule(lr){6-9}\cmidrule(lr){10-12}
& & \textsc{Read}. & \textsc{Insh.} & \textsc{Stru.} & \textsc{Vef.} & \textsc{Con.} & \textsc{Cov.} & \textsc{Fid.} & \textsc{Sem.} & \textsc{Acc.} & \textsc{VQA} \\
\midrule

\multicolumn{12}{c}{\textit{Single-Modal, w/o Search}} \\
\midrule
OpenAI o3-mini
& 31.96 & 51.32 & 52.65 & 37.11 & 13.57 & 28.45 & 33.74 & \bestcell{48.35} & 15.47 & 90.00 & 12.60 \\
DeepSeek-V3.2
& 43.71 & 59.06 & 87.82 & 58.16 & 19.29 & 33.34 & 45.48 & 18.77 & 42.19 & 83.85 & 12.88 \\
Kimi K2 (Thinking)
& 36.91 & 55.60 & 77.27 & 47.34 & 17.14 & 23.54 & 24.62 & 27.20 & 42.00 & 90.00 & 9.50 \\
Qwen 3 235B (A22B)
& 36.04 & 57.77 & 85.74 & 54.05 & 17.14 & 35.60 & 45.73 & 22.98 & 20.43 & 53.09 & 4.95 \\

\midrule
\multicolumn{12}{c}{\textit{Multimodal, w/o Search}} \\
\midrule
Qwen 3 VL 235B (A22B)
& 35.08 & 58.47 & 86.48 & 52.21 & 43.57 & 18.34 & 15.25 & 10.68 & 30.58 & 93.52 & 16.98 \\
GPT-4o
& 28.62 & 54.38 & 68.41 & 40.90 & 10.04 & 10.94 & 4.61 & 11.89 & 24.10 & 71.43 & 18.72 \\
GPT-4.1
& 36.95 & 58.29 & 89.04 & 53.00 & 39.29 & 15.90 & 10.06 & 5.61 & 29.66 & 80.56 & 19.92 \\
GPT-4.1 mini
& 34.23 & 57.61 & 83.62 & 49.60 & 12.86 & 24.20 & 25.44 & 12.33 & 32.62 & 89.91 & 13.21 \\
GPT-4.1 nano
& 28.07 & 51.74 & 64.82 & 37.28 & 10.79 & 18.99 & 19.86 & 24.42 & 27.02 & 76.30 & 13.04 \\
GPT-5 mini
& 38.49 & 51.57 & 81.73 & 47.18 & 39.29 & 20.02 & 26.64 & 32.61 & 33.90 & \bestcell{94.23} & 15.60 \\
GPT-5.1
& 32.69 & 56.41 & 83.57 & 51.80 & 35.71 & 13.22 & 2.30 & 13.67 & 22.03 & 84.29 & 14.32 \\
GPT-5.2
& 32.76 & 59.34 & 83.92 & 54.31 & 46.43 & 14.00 & 1.43 & 5.30 & 12.83 & 50.00 & 9.16 \\
GPT-5.4
& 40.84 & 59.17 & 76.44 & 49.26 & \bestcell{52.45} & 14.56 & 6.75 & 5.75 & 19.56 & 69.23 & 12.60 \\
Grok-3
& 29.89 & 56.14 & 86.13 & 52.24 & 20.00 & 12.57 & 5.79 & 2.80 & 22.18 & 68.39 & 13.89 \\
Grok-4 (Fast Reasoning)
& 36.10 & 57.48 & 80.49 & 52.99 & 36.43 & 17.30 & 14.62 & 6.12 & 28.46 & 87.45 & 19.34 \\

\midrule
\multicolumn{12}{c}{\textit{Multimodal, w/ Search}} \\
\midrule
Claude 4.5 Haiku
& 33.67 & 58.34 & 81.80 & 53.22 & 28.57 & 17.90 & 14.10 & 18.56 & 25.98 & 76.90 & 11.70 \\
Claude 4.5 Sonnet
& 33.61 & 57.14 & 82.31 & 51.65 & 32.14 & 14.36 & 15.09 & 16.11 & 20.73 & 70.13 & 14.41 \\
Claude 4.5 Opus
& 33.84 & 55.73 & 83.86 & 50.70 & 35.00 & 30.64 & 41.14 & 21.97 & 21.30 & 77.21 & 14.75 \\
Gemini 2.5 Flash
& 38.40 & 55.44 & 68.58 & 46.91 & 32.86 & 25.35 & 27.77 & 38.30 & 40.67 & 75.96 & 25.49 \\
Gemini 2.5 Pro
& 38.04 & 56.06 & 85.94 & 51.44 & 38.57 & 30.18 & 28.77 & 14.98 & 19.47 & 92.86 & 12.50 \\
Gemini 3 Flash
& 44.43 & 56.59 & \bestcell{90.22} & 52.00 & 45.71 & 31.95 & 35.07 & 15.42 & 36.61 & 87.31 & 18.99 \\
Gemini 3 Pro
& 44.68 & 57.38 & 75.39 & 49.85 & 46.43 & 37.98 & 41.85 & 6.46 & 40.69 & 80.44 & 23.15 \\
Gemini 3.1 Flash Lite
& 49.08 & 56.56 & 86.11 & 48.88 & 30.89 & 12.28 & 11.65 & 18.52 & \bestcell{53.04} & 87.86 & 28.91 \\
Gemini 3.1 Pro
& \bestcell{50.59} & 59.49 & 87.59 & 52.82 & 35.01 & 13.99 & 12.49 & 11.54 & 46.54 & 83.53 & \multicolumn{1}{>{\columncolor{bestblue}[\tabcolsep][0pt]}c@{}}{\textbf{31.54}} \\

\midrule
\multicolumn{12}{c}{\textit{Deep Research Agent}} \\
\midrule
Tongyi Deep Research (30B-A3B)
& 29.02 & 54.62 & 62.67 & 40.07 & 12.86 & 25.99 & 30.87 & 24.25 & 20.39 & 93.33 & 7.52 \\
Perplexity Sonar Deep Research
& 37.55 & 53.64 & 64.35 & 47.80 & 27.86 & 33.12 & 41.51 & 16.68 & 50.79 & 87.75 & 21.22 \\
ChatGPT Deep Research (o3-mini)
& 29.50 & 53.72 & 63.61 & 37.30 & 29.29 & 10.19 & 4.16 & 11.07 & 27.32 & 73.44 & 21.75 \\
Gemini Deep Research (Gemini 3 Pro)
& 49.41 & \bestcell{75.19} & 89.56 & \bestcell{70.86}
& 35.71 & \bestcell{56.17} & \bestcell{52.84} & 31.29
& 41.29 & 87.54 & 28.45 \\

\bottomrule
\end{tabular}%
}

\end{table}

Table~\ref{main} reports Overall and selected diagnostics. Gemini 3.1 Pro achieves the highest Overall score (50.59), followed by Gemini Deep Research (49.41). Gemini Deep Research leads several displayed FLAE/TRACE dimensions; GPT-5 mini has the highest MOSAIC accuracy score. GPT-5.4 has the largest displayed VEF value (52.45).

\begin{figure}[t]
  \centering
  \includegraphics[width=0.68\textwidth]{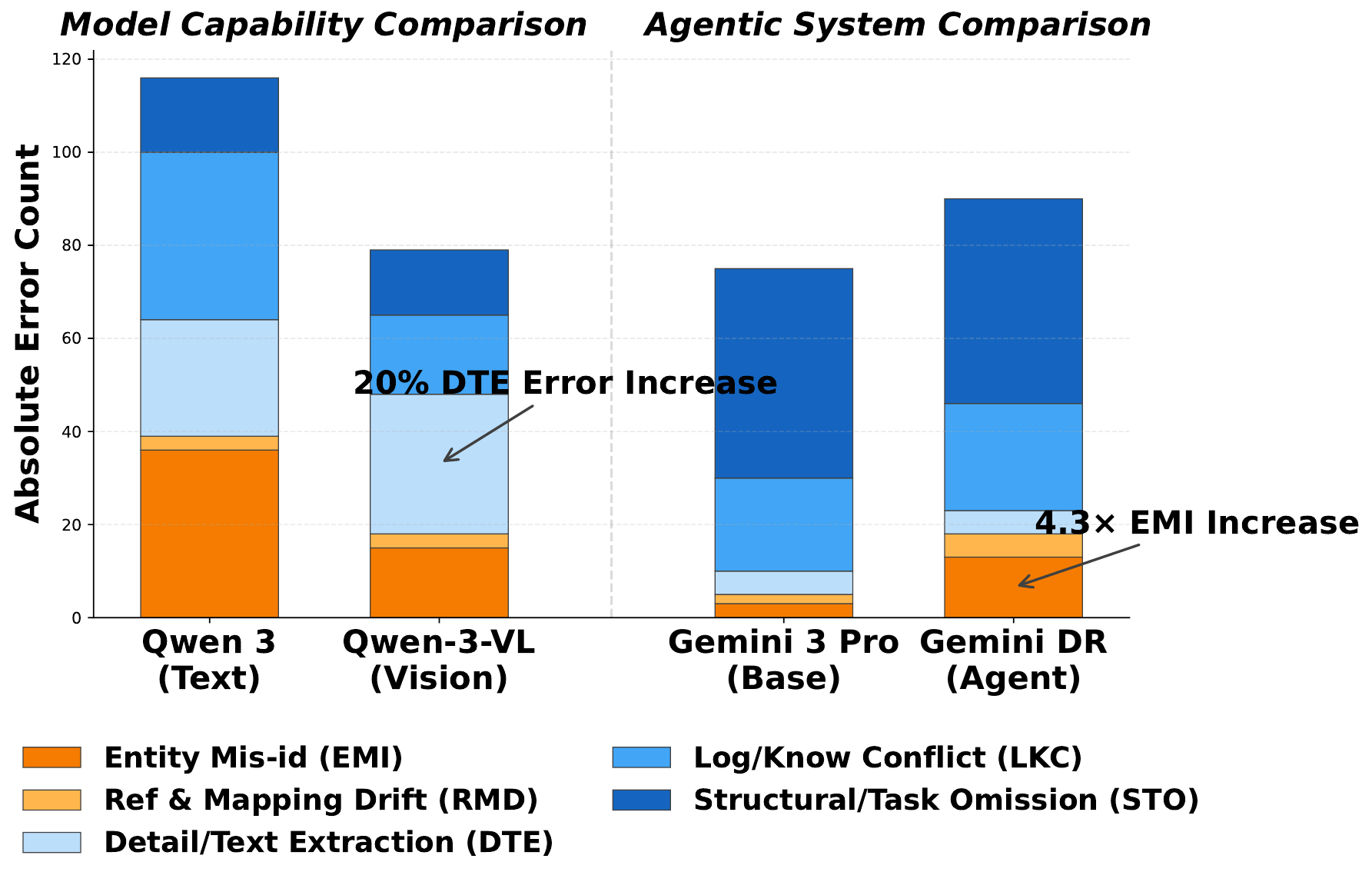}
  \caption{Visual-evidence failure modes for text-only versus vision-enabled backbones (left) and base versus agentic systems (right).}
  \label{fig:unified-error}
\end{figure}

\noindent \textbf{Finding 1: Vision is beneficial only when it is reliable as evidence.} Comparisons within the same model family such as Qwen 3 235B (A22B) vs.\ Qwen 3 VL 235B (A22B) show that adding vision is not a monotonic win. Although their multimodal variants improve visual grounding, the unified error analysis in Figure~\ref{fig:unified-error} reveals increased detail-level extraction failures from mis-reading fine-grained literals such as numerals, dates, labels and table cells. The comparison illustrates that stronger visual alignment can coexist with less reliable extraction of task-critical details. The detailed failure case analysis is in Appendix~\ref{failure case}.

\noindent \textbf{Finding 2: Multimodal alignment and citation grounding can diverge.} Stronger multimodal alignment or prompt-following does not guarantee more reliable citation grounding. Compared with Gemini~3~Pro, its backbone, Gemini Deep Research has higher citation coverage, yet Figure~\ref{fig:unified-error} shows more entity-level failures in its final reports. This coexistence indicates that broader evidence coverage need not ensure correct entity attribution.

\noindent \textbf{Finding 3: Agentic scaffolding helps, but base-model progress can outweigh it.} Gemini Deep Research (Gemini~3 Pro) scores 49.41 versus 44.68 for Gemini~3 Pro, with particularly large gains in citation consistency and coverage. The later Gemini~3.1 Pro reaches 50.59, surpassing the previous-generation deep-research agent.

These results highlight the importance of both agent design and continued backbone improvement: a stronger base model can exceed an agent built on an earlier generation. Tongyi Deep Research (30B-A3B) also scores below several larger systems, and some offline models outperform web-enabled systems on citation coverage, showing that effective evidence use depends on more than tool availability alone.
\par

\subsection{Fine-Grained Domains Analysis}
\label{sec:fine-grained-domains}

\begin{figure}[t]
  \centering
  \includegraphics[width=0.98\textwidth]{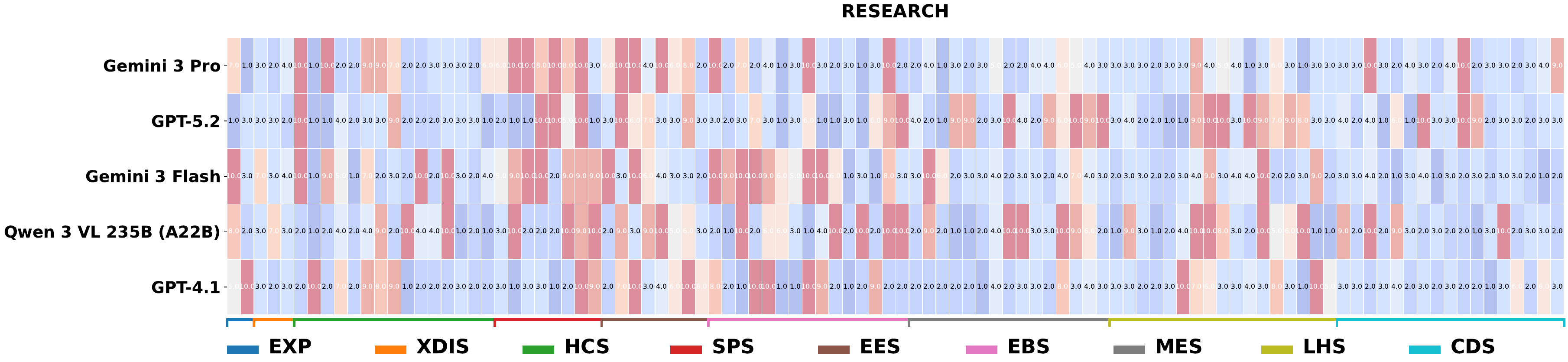}
  \caption{An illustrative view of Research-task scores for five selected systems, grouped by domain. Higher scores indicate better performance.}
  \label{fig:domain-heatmap}
\end{figure}

Figure~\ref{fig:domain-heatmap} illustrates Research-task scores for five selected systems, grouped by domain. Substantial variation within each system shows how a single overall score can conceal task-specific weaknesses. This view complements the aggregate comparison by highlighting the uneven performance of current systems across research tasks.

\subsection{Human Consistency Check}
\label{sec:human-consistency}

We assess evaluator--human agreement on reports from all 140 tasks using pairwise preference agreement (PAR) and Pearson correlation of system-level mean scores (OPC). The annotation protocol, annotator composition, and sampling procedures are detailed in Appendix~\ref{app:human-protocol}.

\begin{table}[!htb]
\caption{Evaluator--human agreement on MMDR-Bench. PAR: agreement with majority human preferences; OPC: Pearson correlation of system-level average scores.}
\label{tab:human-consistency}
\centering
\small
\setlength{\tabcolsep}{4pt}
\renewcommand{\arraystretch}{1.08}
\begin{tabular}{lcc}
\toprule
Method & PAR$\uparrow$ & OPC$\uparrow$ \\
\midrule
Vanilla Prompt Judge & 61.2 & 93.0 \\
\textbf{MMDR-Bench-Eval (Full)} & \textbf{73.5} & \textbf{96.4} \\
w/o \textsc{Vef.} & 68.0 & 95.2 \\
w/o \textsc{MOSAIC} & 70.1 & 95.8 \\
\midrule
Human Inter-Annotator Agreement & 69.8 & -- \\
\bottomrule
\end{tabular}

\end{table}

Table~\ref{tab:human-consistency} shows that the full evaluator achieves 73.5\% PAR and 96.4\% OPC, compared with 61.2\% and 93.0\% for the vanilla prompt judge. Removing \textsc{Vef.} or \textsc{MOSAIC} reduces agreement, supporting their contribution to evaluating multimodal research reports. Additional blinded calibration checks cover report preferences and visual judgments; the two-rater \textsc{MOSAIC} audit yields quadratic-weighted $\kappa=0.887$. Full results and protocols are provided in Appendix~\ref{app:human-protocol}.
\par

\subsection{Robustness of Judge Model and Weights}
\label{sec:robustness-main}

\noindent \textbf{Cross-judge scoring.}
We rescore a fixed set of Gemini-generated reports with Gemini-2.5-Pro and GPT-5.2 under the same parsing, retrieval, and aggregation pipeline. Table~\ref{tab:judge-robustness} (left) shows module-level differences: GPT-5.2 gives lower FLAE and VEF scores and higher textual TRACE scores, while MOSAIC remains close (29.44 vs.\ 29.53). Opposing module shifts yield aggregate means of 36.76 and 37.06, a difference of 0.30 points (about 0.8\%).

\begin{table}[!ht]
\caption{\textbf{Left:} Mean scores under two judge backbones on the same report set. \textbf{Right:} Repeated-scoring ICC(3,1) with Gemini-2.5-Pro as judge; report generators are Gemini-3.1-Pro (Gemini) and GPT-5.4 (GPT). Full consistency statistics appear in Appendix~\ref{app:judge-consistency}.}
\label{tab:judge-robustness}
\centering
\small
\setlength{\tabcolsep}{3pt}
\renewcommand{\arraystretch}{1.08}
\begin{minipage}[t]{0.51\linewidth}
\centering
\begin{tabular}{lcc}
\toprule
Metric & Gem-2.5-Pro & GPT-5.2 \\
\midrule
\textsc{FLAE} & 61.89 & 45.82 \\
TRACE (w/o \textsc{Vef.}) & 28.39 & 39.87 \\
\textsc{Vef.} & 38.57 & 26.42 \\
\textsc{MOSAIC} & 29.44 & 29.53 \\
\midrule
MMDR (w/ \textsc{Vef.}) & 36.76 & 37.06 \\
MMDR (w/o \textsc{Vef.}) & 38.04 & 35.23 \\
\bottomrule
\end{tabular}
\end{minipage}\hfill
\begin{minipage}[t]{0.46\linewidth}
\centering
\begin{tabular}{lcc}
\toprule
Module & Gemini & GPT \\
\midrule
\textsc{FLAE} & 0.295 & 0.153 \\
TRACE (w/o \textsc{Vef.}) & 0.884 & 0.702 \\
\textsc{Vef.} & 0.273 & 0.506 \\
\textsc{MOSAIC} & 0.970 & 0.983 \\
\midrule
Overall MMDR & 0.709 & 0.614 \\
\bottomrule
\end{tabular}
\end{minipage}

\end{table}

\noindent \textbf{Controlled cross-judge rankings.}
We further hold reports, evidence artifacts, prompts, routing, and aggregation fixed on 120 byte-identical reports from six systems and twenty shared tasks (14 Research, 6 Daily). Gemini-2.5-Pro, GPT-5.4, and Grok-4.3 preserve the same winner, with pairwise Spearman correlations of 0.714--0.943. A 10,000-replicate task-paired bootstrap preserves the winner in 98.4\%, 93.3\%, and 99.8\% of resamples, respectively. These results characterize ranking stability on the matched subset; middle-rank intervals are wider. Selection criteria and full uncertainty estimates are in Appendix~\ref{sec:robustness-judge-weights}.
\par

\noindent \textbf{Aggregation-weight sensitivity.}
Six local perturbations around $(w_F,w_T,w_M)=(0.2,0.5,0.3)$ preserve the winner and Top-3, with mean Spearman correlation 0.978 against the official ranking. This supplementary analysis anchors each system at its historical Overall score and applies reweighting deltas from a later independently restarted component run. Separately, 2,000 perturbations of task-adaptive FLAE dimension weights by up to $\pm30\%$ yield mean/minimum Spearman correlations of 0.999/0.995. The component-run provenance, estimated-component exclusion check, and alternative aggregation results are detailed in Appendix~\ref{sec:robustness-judge-weights}.

\noindent \textbf{Repeated-judge consistency.}
We score each fixed report five times with Gemini-2.5-Pro (default temperature 0.2; VQA 0.0). Gemini-3.1-Pro reports cover 18 tasks with three generations each, yielding 270 ratings; GPT-5.4 reports cover 17 tasks with three generations each, yielding 255 ratings. Table~\ref{tab:judge-robustness} (right) reports overall ICC(3,1) of 0.709 and 0.614. MOSAIC is particularly repeatable (0.970/0.983), while FLAE and VEF show greater variability. Binary VEF judgments on the Gemini reports have 73.3\% mean pairwise agreement. Appendix~\ref{app:judge-consistency} gives the complete protocol, Krippendorff's $\alpha$, and coefficients of variation.
\par

\section{Conclusion}
In this work we introduce MMDR-Bench, a benchmark for multimodal evidence use in long-form research reports. Building on the benchmark dataset comprising 140 tasks across 19 domains under Daily and Research regimes, we further propose a unified evaluation pipeline that jointly measures report quality, citation-grounded faithfulness, and text--visual evidence consistency. Results across 28 state-of-the-art LLMs and DRAs reveal persistent trade-offs between writing quality, citation discipline, and multimodal grounding.

\section{Limitations}
\label{sec:limitations}
MMDR-Bench currently contains 140 tasks, predominantly in English and Chinese. Its evaluation focuses on groundedness in retrieved sources and visual evidence, with coverage and score stability affected by inaccessible sources, changing hosted models, and judge variability. Broader language coverage and continued benchmark maintenance are important directions for future work. Further methodological and implementation details are discussed in Appendix~\ref{app:limitations}.

\subsection*{Broader Impact and Ethics Statement}
Human involvement comprises researcher-led task construction, author-group report assessment, and supplementary blinded audits with non-author raters (Appendix~\ref{app:human-protocol}); results are reported in aggregate. Task images and cited materials include third-party content, whose reuse remains subject to the respective source terms and licenses. Health- and finance-related reports are evaluation artifacts and require appropriate expert review before use in consequential decisions.

\subsection*{Reproducibility statement}
The dataset and evaluation-code links are provided in the abstract. Dataset construction and visual ground-truth preparation are described in Appendix~\ref{app:dataset-details}; evaluator formulas, weights, prompts, and fallback behavior are documented in Appendices~\ref{app:metrics} and~\ref{app:flae}. Model access information, evaluation coverage, human assessment, and robustness protocols appear in Appendices~\ref{app:model_details}, \ref{app:reliability}, \ref{app:human-protocol}, and~\ref{sec:robustness-judge-weights}, respectively. We distinguish frozen-report rescoring from regeneration through hosted APIs. The availability of backend snapshot identifiers and original human annotation records is documented alongside the relevant protocols.

\bibliography{reference}
\bibliographystyle{unsrtnat}

\appendix

\section{Appendix}

\subsection{Additional Dataset Construction Details}
\label{app:dataset-details}

\noindent \textbf{Two Task Regimes.}
MMDR-Bench contains 140 researcher-curated tasks spanning 19 domains, which are organized into two complementary regimes:
Daily (40 tasks across 10 domains) and Research (100 tasks across 9 domains).
Daily tasks reflect everyday deep-research workflows driven by loosely structured visual inputs,
such as screenshots, photographs, and user-facing interfaces, and require lightweight yet verifiable evidence use.
In contrast, Research tasks span scientific and social-science domains and emphasize analysis-heavy settings
with information-dense visuals, including charts, diagrams, and tables, where models must integrate
multi-source evidence and synthesize it into coherent, citation-grounded reports under stricter fidelity constraints.

\noindent \textbf{Researcher-Driven Task Proposal and Refinement.}
Researchers with relevant disciplinary training propose candidate tasks, which we iteratively refine with clarity checks, multimodal-necessity checks (tasks must be image-dependent), and evidence-grounding checks (reports must be verifiable via citations).
This process yields the final 140-task benchmark. The source pool contains 98k+ real-world queries and supplies material for task curation. The pool size describes the source collection rather than a selection ratio. The benchmark pairs 140 curated tasks with multi-stage evidence assessment, averaging 35.48 evaluation calls per report (Appendix~\ref{app:cost}); task-paired bootstrap estimates for the matched study are reported in Appendix~\ref{sec:robustness-judge-weights}.

\noindent \textbf{Multimodal Packaging, Difficulty, and Multilinguality.}
Each task is packaged as an image--text bundle with a variable number of images.
We annotate instance difficulty (easy, hard, complex) and record the task language.
The benchmark is multilingual, dominated by English and Chinese, with additional languages in the long tail.

\subsubsection{Standardized Report Generation Protocol}
\label{app:report-rules}

We prompt models to generate reports grounded in Claim--URL verifiability and inline visual support.
Citations should appear immediately after factual claims, and each citation index must map to exactly one URL in a \texttt{References:} block.
For image-dependent claims, the report should embed the referenced image inline with a short caption carrying the same citation index, and task-provided input images should be embedded before drawing image-dependent conclusions.
We encourage diverse sources (Daily about 6 and Research about 10 when feasible) while avoiding social media and question-answering forums, and we note that Daily inputs can be noisy or partial.

\subsubsection{Textualized Visual Ground Truth for VEF}
\label{app:vef-gt}
\textsc{Vef.} uses a task-specific textualized visual ground truth (visual GT) that records directly observable facts from the task-provided images.
Concretely, when a task author prepares a benchmark task, they additionally write a concise visual GT by describing the key visual facts in the image bundle as-is.
Depending on the visual type, the GT may include: salient entities/objects and their identities; explicit numbers, labels, and table entries; chart titles, axes, units, legends, and trend/ordering relations; and screenshot/UI states such as selected options, warnings, and on-screen text.
This GT is intended to be minimal yet sufficient to distinguish correct visual interpretation from visually grounded hallucinations such as mis-identifying the central entity in an image.
\par

\noindent \textbf{Decision Rule and Threshold.}
The judge returns a 0--10 score and a raw PASS/FAIL label. In the released worker, the numerical endpoint is $100\,\mathbf{1}[s_{\mathrm{judge}}\geq6]$; the numerical score determines the endpoint, and the raw PASS/FAIL label is retained for diagnostics. The judging instructions classify wrong identities and false presence as failures. The binary endpoint gives critical visual errors an explicit penalty within VEF, while the 0--10 score and raw label provide diagnostic detail. Historical score provenance is documented in Appendix~\ref{app:exp-details}.

\noindent \textbf{Future Versioning and Drift Management.}
Building on retained task/GT hashes for the audited subsets, future maintenance will extend immutable version identifiers, release snapshots, correction records, and canary regression checks to the benchmark.

\noindent \textbf{VEF Score Semantics.}
For the set $\mathcal D_s$ with retained final VEF outcomes, let $y(t,R)\in\{0,1\}$ denote that binary outcome. We report
\begin{equation}
\textsc{Vef.}(\mathcal D_s)=\frac{100}{|\mathcal D_s|}\sum_{t\in\mathcal D_s}y(t,R).
\end{equation}
This defines the binary endpoint; historical aggregation and coverage are documented in Appendix~\ref{app:exp-details}.

\noindent \textbf{Quality Control and Calibration.}
Task authors inspect the input images to construct the textualized visual GT. As shown in Figure~\ref{fig:trace-vef-prompt}, the VEF judge receives the task text, visual GT, and report, rather than the images themselves. Human spot-checks on a small development subset inform refinements for incorrect entity identity, swapped labels, and incorrect numeric readings. Repeated-judging results appear in Appendix~\ref{app:judge-consistency}.

\subsection{Evaluation Metrics}
\label{app:metrics}

We score each generated report with three modules: \textsc{FLAE} (generation quality), \textsc{TRACE} (evidence and task faithfulness), and \textsc{MOSAIC} (image-grounded integrity).  Figure~\ref{fig:Architecture} summarizes the conceptual dependencies.

\subsubsection{Weighting Hyperparameters}
\label{app:weighting}
Table~\ref{tab:weighting} summarizes the weighting hyperparameters used by the evaluator, including task-adaptive weights and fixed shares.

\begin{table*}[!htbp]
\caption{Weighting configuration. Adaptive dimension weights and fixed channel/module shares are distinct design choices.}
\label{tab:weighting}
\centering
\small
\setlength{\tabcolsep}{5pt}
\renewcommand{\arraystretch}{1.15}
\begin{tabularx}{\textwidth}{@{}l l X@{}}
\toprule
Scope & Rule & Weighted quantities \\
\midrule
Overall & Fixed & FLAE/TRACE/MOSAIC shares of 0.20/0.50/0.30. \\
FLAE dimensions & Task-adaptive & Readability, insightfulness, coherence, and structure; normalized weights $W_d$. \\
FLAE fusion & Type-fixed & Coefficient $\alpha$ combines fixed-transform and holistic scores; see Appendix~\ref{app:flae-fusion-prompts}. \\
TRACE visual fidelity & Fixed & $\lambda_{\textsc{Vef.}}=0.4$, giving VEF an Overall share of 0.20. \\
TRACE evidence & Task-adaptive & Consistency, coverage, fidelity, and source diversity; normalized weights $W_k$. \\
MOSAIC items & Type-fixed & Routed item weights in Table~\ref{tab:mosaic-item-weights}. \\
\bottomrule
\end{tabularx}

\end{table*}

\noindent \textbf{Rationale.}
Task-adaptive dimension weights are intended to accommodate heterogeneous tasks. Fixed module shares encode the benchmark's emphasis on evidence support. Sensitivity checks characterize ranking changes under the specified alternative configurations; the weights are design settings.

\subsection{Module Implementation Details}
\label{app:flae}
\subsubsection{Measured and Judge-Estimated Features}
\label{app:flae-features}
The directly measured features include type--token ratio $TTR$, heading density $H$, and long-sentence ratio $L$. The judge supplies $E$, $p_{\mathrm{analysis}}$, $\rho_{\mathrm{connective}}$, $\rho_{\mathrm{synthesis}}$, $U,O,T$, $\kappa$, $\theta$, and section-presence flags, together with holistic dimension scores. Here $E$ represents errors; $p_{\mathrm{analysis}}$, $\rho_{\mathrm{connective}}$, and $\rho_{\mathrm{synthesis}}$ describe analytical/synthetic content; $U,O,T$ are coherence sub-scores; $\kappa$ is the conflict term; and $\theta$ is the structural-adequacy factor. Insightfulness operationalizes analytical and synthetic content through these judge-estimated attributes. Retained feature values permit deterministic recomputation of the transforms.

\subsubsection{Fixed Formula Channel}
\label{app:flae-formulas}
Let $c(x)=\min(1,\max(0,x))$, $\sigma(x)=(1+e^{-x})^{-1}$, and $H_m$ be the harmonic mean with a $10^{-9}$ numerical floor. The retained evaluator computes
\begin{equation}
\label{eq:app-flae-short}
\begin{aligned}
R_{\mathrm{form}}&=c\bigl(\sigma(2TTR+1.2H-2L-3E)\bigr),\\
I_{\mathrm{form}}&=c\left[\left(\max(10^{-9},p_{\mathrm{analysis}})\,
\max(10^{-9},\rho_{\mathrm{connective}})\,
\max(10^{-9},\rho_{\mathrm{synthesis}})\right)^{1/3}\right],\\
C_{\mathrm{form}}&=c\bigl(H_m(U,O,T)\max(0,1-2\kappa)\bigr),\\
S_{\mathrm{form}}&=c\bigl(P_{\mathrm{sections}}\min(1,n_{\mathrm{img}}/n^*_{\mathrm{img}})c(\theta)\bigr).
\end{aligned}
\end{equation}
Here $n_{\mathrm{img}}$ counts unique HTTP(S) Markdown image URLs, and $n^*_{\mathrm{img}}=\max(1,n_{\mathrm{words}}/350)$ for evaluator label \texttt{research}, or $\max(1,n_{\mathrm{words}}/700)$ otherwise. $P_{\mathrm{sections}}$ is the product of the implementation's required-section indicators. The research list comprises Executive Summary, Background, Methodology, Analysis \& Evidence, Visual Integration, and Conclusion; other labels use the first, fourth, and sixth entries.

\noindent \textbf{Evaluator Version and Structural Scoring.}
The reported results use the same evaluator version across systems: its lowercase detected-section keys are looked up with title-case required keys, yielding $P_{\mathrm{sections}}=S_{\mathrm{form}}=0$, so structural scoring retains the weighted judge-channel contribution and the section-based $g_S{+}=1$ adjustment always applies.

\subsubsection{Judge Inputs and Output Schema}
\label{app:flae-judge-prompt}
The feature/scoring call receives the task, evaluator report-type hint, and the first 1,400/14,000 characters of the rules/report, returning strict JSON with the feature fields, section-presence map, and optional $[0,1]$ holistic fields \texttt{llm\_R}, \texttt{llm\_I}, \texttt{llm\_C}, and \texttt{llm\_S}. Missing holistic fields use the corresponding formula scores; failed structured calls use the implementation's default feature/score record.

\subsubsection{Task-Adaptive Weights and Type-Fixed Fusion}
\label{app:flae-fusion-prompts}
\label{app:flae-alpha-ablation}
Fusion uses the retained per-report evaluator label and coefficient; evaluator labels may differ from the benchmark's canonical task regimes. The coefficient follows a fixed lookup:
\begin{equation}
\alpha=\begin{cases}
0.60,&\texttt{research},\\
0.40,&\texttt{daily\_hard},\\
0.30,&\texttt{daily\_easy},\\
0.35,&\text{otherwise}.
\end{cases}
\qquad
s_d=\alpha s_d^{\mathrm{form}}+(1-\alpha)s_d^{\mathrm{judge}}.
\label{eq:flae-alpha-fuse}
\end{equation}
This is distinct from adaptive dimension weighting. A separate task-controller call receives the question and report-type hint and returns four-dimensional adjustment vectors \texttt{v\_general} and \texttt{v\_evidence}. For FLAE,
\begin{equation}
W_d=\operatorname{softmax}_d\bigl(\log(\pi_d+10^{-9})+v_d+g_d\bigr),\qquad \pi=(0.25,0.25,0.25,0.25).
\end{equation}
In $(R,I,C,S)$ order, $g$ starts at zero. The code adds 1 to $g_S$ if fused $S<0.5$ or $P_{\mathrm{sections}}=0$; adds 0.8 to $g_R$ if $E>0.10$ or $L>0.35$; and adds 0.4 to both $g_I$ and $g_C$ if $H>0.12$ and $S>0.7$. These rules define the implemented weighting configuration.

\subsubsection{TRACE: Evidence and Task Faithfulness}
\label{app:trace}

\textsc{TRACE} evaluates citation fidelity and multimodal task faithfulness, aggregating as Eq.~\eqref{eq:trace-score}. We parse citation markers and the references: block to build a one-to-one index-to-URL map, extract atomic claims, and align each claim to its cited URL(s) to form claim-URL pairs. For accessible pages, a judge verifies whether each claim is supported and aggregates citation fidelity into $\mathcal{K}$: \textsc{Con.}, \textsc{Cov.}, \textsc{Fid.}, and \textsc{Div.}. We additionally compute \textsc{Vef.} by matching the report to a textualized visual ground truth for the task, then threshold it at $\tau_{\textsc{Vef.}}$.
\par

\noindent \textbf{Text-Evidence Components and Weights.}
Consistency is a relevance-weighted support mean multiplied by one minus the prohibited-source fraction. Coverage is a soft supported fraction over extracted key textual claims, with the denominator restricted to extracted key textual claims and image-dependent claims excluded. Fidelity is the harmonic mean of mean non-hallucination, non-inversion, and chain-soundness scores. Diversity is normalized domain entropy multiplied by $\min(1,(m_{\mathrm{auth}}+0.5(m_{\mathrm{acc}}-m_{\mathrm{auth}}))/3)$, where the counts denote authoritative and non-prohibited domains.
\par

Evidence weights are $W=\operatorname{softmax}(\log(\pi+10^{-9})+v+h)$ with prior $\pi=(0.40,0.25,0.25,0.10)$ in (Con., Cov., Fid., Div.) order and task-controller vector $v$. Starting from zero, the diagnostic adjustment $h$ adds 1 to Con. and subtracts 0.8 from Div. when the prohibited-source fraction exceeds 0.05; adds 0.8 to Cov. when at least three key claims have coverage below 0.4; and adds 0.8 to Fid. when mean hallucination exceeds 0.25 or inversion exceeds 0.15. Empty text-evidence sets receive a zero aggregate. For non-prohibited sources whose fetch fails, the retained implementation floors support at 0.18 (excluding direct-image/local items); this heuristic fallback distinguishes an unverified citation from a contradicted one without assigning full support.

\noindent \textbf{VEF Thresholding.}
The threshold and diagnostic fields are defined in Appendix~\ref{app:vef-gt}; repeated-judging consistency is reported in Appendix~\ref{app:judge-consistency}.

\noindent \textbf{VEF Judge Prompt Template.}
Figure~\ref{fig:trace-vef-prompt} shows the prompt used to score \textsc{Vef.} against the task-specific textualized visual ground truth and to output a PASS/FAIL verdict.
\begin{figure*}[t]
\fbox{
\begin{minipage}{0.97\textwidth}
\footnotesize
\textbf{SYSTEM}\\
You are a STRICT QA Judge for \textsc{Vef.}. Use the task, the provided visual ground truth, and the report.\\
Return \textbf{only} valid JSON.

\textbf{USER}\\
\textbf{Segment:}\\
\{SEGMENT\}\\
\textbf{Question:}\\
\{TASK\_TEXT\}\\
\textbf{Visual ground truth (text-form requirements):}\\
\{VEF\_GT\}\\
\textbf{Report:}\\
\{REPORT\_TEXT\}\\

Rules: any wrong visual identity is FAIL; any false presence is FAIL; missing details allowed only if no wrong identities; score below 6 must be FAIL.\\
Output JSON with keys: score, reason, verdict (PASS or FAIL).\\
\end{minipage}
}
\caption{TRACE prompt template for \textsc{Vef.}.}
\label{fig:trace-vef-prompt}
\end{figure*}

\subsubsection{MOSAIC: Image-Grounded Integrity}
\label{app:mosaic}

\textsc{MOSAIC} checks whether image-grounded statements match the referenced visuals and aggregates by Eq.~\eqref{eq:mosaic-item-score}. It extracts figure-linked items and routes them into visual types (photo, datachart, ocrchart, diagram). Each item is scored with a formatting and integration factor $f_i\in[0,1]$; for diagrams we additionally use a hallucination factor $h_i\in[0,1]$.

\begin{table}[t]
\caption{Default \textsc{MOSAIC} item-score weights by routed visual type.}
\label{tab:mosaic-item-weights}
\centering
\small
\setlength{\tabcolsep}{5pt}
\renewcommand{\arraystretch}{1.10}
\begin{tabularx}{0.55\columnwidth}{@{}l >{\raggedright\arraybackslash}X@{}}
\toprule
Visual type &
\makecell[l]{Item score $s_i$\\(normalized)} \\
\midrule
\textsc{datachart} / \textsc{ocrchart} &
$s_i=f_i\cdot \bigl(0.9\,\textsc{Acc.}_i + 0.1\,\textsc{VQA}_i\bigr)$ \\
\textsc{photo} &
$s_i=f_i\cdot \bigl(0.5\,\textsc{Sem.}_i + 0.5\,\textsc{VQA}_i\bigr)$ \\
\textsc{diagram} &
$s_i=f_i\cdot \bigl(0.5\,\textsc{VQA}_i + 0.5\,(1-h_i)\bigr)$ \\
\bottomrule
\end{tabularx}

\end{table}

\noindent \textbf{Visual Claim Support and Temperature.}
The Gemini VQA module asks whether the image supports its associated statement at temperature 0.0 (other primary-judge calls: 0.2), using up to five first-token log-probability candidates to compute positive mass (yes/true/correct) divided by positive plus negative mass (no/false/incorrect/not). It falls back to returned yes/no text when log-probabilities are unavailable, and to zero for insufficient recognized mass or API exceptions.
\par

\subsection{Evaluated Model Details}
\label{app:model_details}
Table~\ref{tab:model_details} lists the 28 evaluated systems, access platforms or model families, usage settings, and testing windows. The original systems were tested in December 2025; GPT-5.4, Gemini 3.1 Pro, and Gemini 3.1 Flash-Lite were added in April 2026, using Azure for GPT-5.4 and the Gemini API for the two Gemini models. Exact backend snapshot identifiers are controlled by the serving platforms and are unavailable in our retained records. We report model names, access platforms, and testing windows as the available identifiers. Exact regeneration is constrained by platform-managed backend changes; frozen-report rescoring holds the generated reports fixed. The retained label ``ChatGPT Deep Research (o3-mini)'' denotes the historical Azure-based setup; its provenance is recorded at the platform and deployment-label level. We group systems into three tiers to reflect progressively stronger tool access and orchestration capability. Tier~1 covers single-shot LLM/LMM baselines with external browsing disabled, corresponding to the first two groups in Table~\ref{main}. For the text-only baselines (o3-mini, DeepSeek-V3.2, Kimi K2, and Qwen 3 235B), retained task inputs contain query text and textual image-file references; these references do not convey image content, and the systems serve as no-vision reference points. Tier~2 includes web-enabled report generators with built-in browsing, representing mainstream tool-using LMM deployments. Tier~3 contains dedicated Deep Research Agents that explicitly orchestrate multi-step retrieval and synthesis, and thus reflect agentic search-and-write behavior beyond a single response.

\begin{table*}[!b]
\caption{Evaluated systems, access information, and testing windows. Identifiers comprise model names, access platforms, and testing windows; exact platform-managed backend snapshots are unavailable.}
\label{tab:model_details}
\centering
\scriptsize
\setlength{\tabcolsep}{4pt}
\renewcommand{\arraystretch}{1.2}
\begin{threeparttable}

\begin{tabularx}{\textwidth}{@{}l l l l X@{}}
\toprule
Tier & System (paper) & Modality/Setting & Testing window & Provider / access \\
\midrule
\multicolumn{5}{l}{\textbf{Tier 1: Single-shot LLM/LMM baselines}} \\
T1 & OpenAI o3-mini & Text/Reasoning & December 2025 & OpenAI models\tnote{a} \\
T1 & DeepSeek-V3.2 (Base) & Text & December 2025 & DeepSeek API\tnote{b} \\
T1 & Kimi K2 (Thinking) & Text/Reasoning & December 2025 & Kimi API\tnote{c} \\
T1 & Qwen 3 235B (A22B) & Text & December 2025 & Qwen API\tnote{d} \\
T1 & Qwen 3 VL 235B (A22B) & Multimodal & December 2025 & Qwen-VL\tnote{d} \\
T1 & GPT-4o & Multimodal & December 2025 & OpenAI models\tnote{a} \\
T1 & GPT-4.1 & Multimodal & December 2025 & OpenAI models\tnote{a} \\
T1 & GPT-4.1 mini & Multimodal & December 2025 & OpenAI models\tnote{a} \\
T1 & GPT-4.1 nano & Multimodal & December 2025 & OpenAI models\tnote{a} \\
T1 & GPT-5 mini & Multimodal & December 2025 & OpenAI models\tnote{a} \\
T1 & GPT-5.1 & Multimodal & December 2025 & OpenAI models\tnote{a} \\
T1 & GPT-5.2 & Multimodal & December 2025 & OpenAI models\tnote{a} \\
T1 & GPT-5.4 & Multimodal & April 2026 & Azure \\
T1 & Grok-3 & No web browsing & December 2025 & xAI API\tnote{e} \\
T1 & Grok-4 (Fast Reasoning) & Reasoning & December 2025 & xAI API\tnote{e} \\
\addlinespace

\multicolumn{5}{l}{\textbf{Tier 2: Web-enabled report generators}} \\
T2 & Claude 4.5 Haiku & Web-enabled & December 2025 & Anthropic\tnote{f} \\
T2 & Claude 4.5 Sonnet & Web-enabled & December 2025 & Anthropic\tnote{f} \\
T2 & Claude 4.5 Opus & Web-enabled & December 2025 & Anthropic\tnote{f} \\
T2 & Gemini 2.5 Flash & Web-enabled & December 2025 & Gemini API\tnote{g} \\
T2 & Gemini 2.5 Pro & Web-enabled & December 2025 & Gemini API\tnote{g} \\
T2 & Gemini 3 Flash & Web-enabled & December 2025 & Gemini API\tnote{g} \\
T2 & Gemini 3 Pro & Web-enabled & December 2025 & Gemini API\tnote{g} \\
T2 & Gemini 3.1 Pro & Web-enabled & April 2026 & Gemini API \\
T2 & Gemini 3.1 Flash-Lite & Web-enabled & April 2026 & Gemini API \\
\addlinespace

\multicolumn{5}{l}{\textbf{Tier 3: Deep Research Agents (DRA)}} \\
T3 & Gemini Deep Research (Gemini 3 Pro) & Agentic DR & December 2025 & Google DR\tnote{h} \\
T3 & ChatGPT Deep Research (o3-mini) & Agentic DR & December 2025 & Azure (historical label) \\
T3 & Tongyi Deep Research (30B-A3B) & Agentic/IR & December 2025 & Tongyi DR\tnote{i} \\
T3 & Perplexity Sonar Deep Research & Agentic/IR & December 2025 & Perplexity\tnote{j} \\
\bottomrule
\end{tabularx}

\begin{tablenotes}[flushleft]
\scriptsize
\item[a] OpenAI model-family and system documentation: \citep{openai_o3_o4_mini_system_card,openai_gpt5_system_card,gpt4}.
\item[b] DeepSeek documentation: \citep{deepseekv}.
\item[c] Kimi documentation: \citep{kimi}.
\item[d] Qwen/Qwen-VL documentation: \citep{qwen3,Qwen3-VL}.
\item[e] xAI Grok documentation: \citep{xai_grok}.
\item[f] Anthropic Claude documentation: \citep{anthropic_claude}.
\item[g] Gemini model-family documentation: \citep{gemini_family}.
\item[h] Google Gemini Deep Research: \citep{google_gemini_deep_research,gemini_family}.
\item[i] Tongyi Deep Research: \citep{tongyidr}.
\item[j] Perplexity Sonar Deep Research: \citep{perplexity_ai}.
\end{tablenotes}
\end{threeparttable}

\end{table*}

\subsection{Coverage, Reliability, and Missing Data}
\label{app:reliability}

\subsubsection{Additional Experimental Setup and Reliability Handling}
\label{app:exp-details}

We report reliability-related settings that affect score stability and effective coverage.

\noindent \textbf{Primary Judge Configuration.}
Gemini-2.5-Pro is the primary judge. General judging calls default to temperature 0.2, while the VQA yes/no module uses 0.0 to reduce sampling variability. Its token-level scoring and fallbacks are specified in Appendix~\ref{app:mosaic}.

\noindent \textbf{TRACE Weighting.}
With the benchmark-level module weight $w_T=0.5$, we set $\lambda_{\textsc{Vef.}}=0.4$ so that \textsc{Vef.} contributes $0.5\times 0.4=0.2$ of the overall score. The remaining $0.3$ overall share from \textsc{TRACE} is allocated to citation-fidelity metrics (\textsc{Con.}, \textsc{Cov.}, \textsc{Fid.}, \textsc{Div.}) using task-adaptive weights.
\par

\noindent \textbf{MOSAIC Gate and Missing Outputs.}
MOSAIC is gated on valid positive report-quality and text-evidence scores. A missing or unusable MOSAIC output contributes zero within a scored task; this differs from excluding a task for which no complete pipeline score is available. No reason-aware validity penalty is applied to Table~\ref{main}.
\par

\noindent \textbf{Score Provenance and Coverage.}
Table~\ref{main} preserves the original reported values. Historical VEF verdicts were also evaluated separately and incorporated offline. Reconciliation of the main-table values with separately retained run-level aggregation records and per-task denominator manifests remains incomplete; results from distinct runs are kept separate. The benchmark contains 140 tasks, with repeated provider refusals for some image-based requests. MOSAIC submetrics are conditional means over applicable task-level dimension summaries with retained scores, so denominators can differ across systems. The later-run components and estimated values used for weight sensitivity are identified in Appendix~\ref{sec:robustness-judge-weights}.
\par

\noindent \textbf{Interpretation of Tables and Domain Plots.}
Column highlights in Table~\ref{main} mark descriptive maxima among the displayed means. Figure~\ref{fig:domain-heatmap} illustrates actual Research-task scores for five selected systems across unequal-sized domains. Cross-generation comparisons involve the evaluated backbones, tool configurations, testing windows, and effective coverage.

\subsubsection{Evaluation Cost and Timing}
\label{app:cost}

A full end-to-end evaluation from raw reports requires 35.48 model calls per report on average (median 35, p90 56, max 70). The mean call decomposition is 2.00 calls for \textsc{FLAE}, 14.92 for \textsc{TRACE}, 1.00 for \textsc{Vef.}, and 17.56 for \textsc{MOSAIC}. Evaluating one system on all 140 tasks therefore requires roughly 4,967 evaluation calls; extrapolating gives about 139,000 calls for one 28-system scoring cycle.
\par

These figures cover evaluation only. Across 3,439 reports with retained timing instrumentation, introduced partway through the runs, median end-to-end scoring time was 163 seconds per report (mean 185, p90 356). Under the Gemini-2.5-Pro configuration, scoring cost about USD 0.36 per report, or approximately USD 50 to evaluate an additional system on 140 tasks.

\subsubsection{Failure-Reason Taxonomy}
\label{app:na-weights}
Missing or unscorable outputs may arise from unusable model responses, parser/schema failures, provider refusals or infrastructure errors, and inaccessible evidence assets. These categories support diagnostic interpretation of the retained logs. Provider and access failures indicate operational constraints; reason tags support diagnosis.

\subsubsection{Failure Case Analysis}
\label{failure case}

For qualitative analysis of multimodal task understanding, we select reports marked FAIL by the Visual Evidence Fidelity (\textsc{Vef.}) check in \textsc{TRACE}. We examine visual-evidence issues in this judge-selected report set using the five categories below; other metrics provide complementary diagnostics.

We group \textsc{Vef.}-level failures into five categories: EMI (entity mis-identification), RMD (reference or mapping drift), DTE (detail or symbol extraction errors), LKC (logical or knowledge conflicts), and STO (structural or task-level omissions). Figure~\ref{fig:unified-error} summarizes the distribution of these failure modes, comparing (left) a text-only backbone versus a vision-enabled backbone and (right) a base model versus its agentic deep-research system.

On the left, the vision-enabled model shows a clear increase in DTE, consistent with mis-reading fine-grained literals such as small numerals, axis labels, timestamps, or table cells. These local errors leave some report claims unsupported.

On the right, the agentic system exhibits more EMI failures in final reports, including evidence bound to an incorrect referent. Attributing these final-report errors to intermediate retrieval or reasoning steps requires process logs.

\begin{figure}[!htb]
\centering
\fbox{\begin{minipage}{0.94\textwidth}
\small
\textbf{Retained VEF FAIL example: Qwen3-VL, R0012}\par
\textbf{Task visuals.} A population table, an unemployment plot, and an asset/output plot for Japan.\par
\textbf{Textual visual GT (translated summary).} Image 00: population data for 2000--2015; Image 01: long-term unemployment rates; Image 02: asset/output indices.\par
\textbf{Report headings (translated from Chinese).} ``Image 00: Japan's unemployment rate''; ``Image 01: Assets/output''; ``Image 02: Population changes.''\par
\textbf{Retained judge result.} FAIL; diagnostic score 1/10 under Gemini-2.5-Pro. The report cyclically assigns all three image numbers to the wrong visual content.\par
\textbf{Failure category.} Reference or mapping drift (RMD): statements are attributed to the wrong task images.\par
\textbf{Selection.} This report is retained in the matched audit; the example uses its existing evaluation record.
\end{minipage}}
\caption{A retained VEF failure from Qwen3-VL on task R0012. The visual GT and report headings are translated for presentation; the FAIL label and diagnostic score come from the retained evaluation record.}
\label{fig:vef-failure-example}
\end{figure}

\subsection{Human Evaluation Protocol}
\label{app:human-protocol}

The original author-group study assesses evaluator--human agreement; two supplementary blinded audits below examine report preferences and visual judgments on frozen artifacts.
We use the full 140 tasks from both Daily and Research regimes and collect reports from all evaluated models.
For each task, we form balanced system pairs by stratifying on overall score and tier (to avoid trivial comparisons), and sample a fixed number of pairs per task.
Twelve annotators with relevant disciplinary background independently assess report pairs: for each pair, three annotators provide an overall preference and a coarse score on the same rubric, and we aggregate by majority vote.
Evaluator--human agreement is measured by pairwise agreement (PAR) on preferences and score correlation (OPC), where OPC is the Pearson correlation between system-level mean scores computed by averaging over tasks and sampled pairs.
We also manually spot-check extracted Claim--URL units and supporting snippets in a subset of pairs. These checks provide a qualitative audit of the sampled evidence units.
Only aggregate results remain available for the original study; exact pair counts, individual votes, tie handling, and the computation underlying the reported 69.8\% inter-annotator agreement cannot be reconstructed from them.

We additionally conduct two small audits on frozen artifacts with samples and endpoints distinct from the original study. In the preference pilot, two non-author raters blindly assess twenty score-independently selected report pairs with system identities and evaluator scores hidden. They agree on 13/20 A/B/Tie labels (Cohen's $\kappa=0.466$); an author who had not previously read the paired reports adjudicates the seven disagreements using the same hidden-identity interface. In this small three-class pilot (A/B/Tie), pooled strict agreement is 40.0\% at the default and 38.3--45.0\% across six local weight settings; the differences are statistically inconclusive.

In the visual audit, two blinded raters assess 48 self-contained \textsc{MOSAIC} visual judgments, yielding human--human quadratic-weighted $\kappa=0.887$. All 13 VQA items classified as supported by \textsc{MOSAIC} receive the maximum human rating. This positive-item agreement concerns the audited evaluator-positive items; the audit does not estimate recall or false-negative rates.

\subsection{Robustness of Judge Model and Weights}
\label{sec:robustness-judge-weights}

We test robustness to judge backbones and aggregation weights in two ways.
First, we retain the cross-judge comparison from the original experiment, where a fixed set of Gemini-generated reports is re-scored by two judge backbones under the same parsing, retrieval, and aggregation pipeline.
Second, to isolate judge choice more strictly, we conduct a controlled same-report study on 120 byte-identical reports from six systems and twenty shared tasks, with reports, evidence artifacts, prompts, routing, score ranges, and aggregation held fixed.

Table~\ref{tab:judge-robustness} (left) shows module-level differences on the fixed report set: GPT-5.2 assigns lower mean \textsc{FLAE} and \textsc{Vef.} scores, while mean \textsc{MOSAIC} scores are similar.

Despite per-stage shifts, the overall MMDR score changes by only 0.30 points (36.76 to 37.06, about 0.8\%), with opposing module shifts contributing to aggregate stability. Repeated-scoring consistency is assessed separately below.

\noindent \textbf{Controlled same-report judge study.}
The controlled subset contains six systems and twenty common tasks (14 Research, 6 Daily), selected before new judging using only exact artifact availability, provider/model-family diversity, rank strata, system category, and task metadata. Rank stratification used previously available scores, and selection was finalized before new cross-judge scores and disagreement outcomes were computed.
We compare retained exact-match Gemini-2.5-Pro evaluations with new GPT-5.4 and Grok-4.3 rescoring. All three judge backbones preserve the winner. Pairwise system-ranking correlations are descriptive because $n=6$, and characterize ordering across providers: Spearman/Kendall are 0.714/0.467 for Gemini vs.\ GPT-5.4, 0.943/0.867 for Gemini vs.\ Grok-4.3, and 0.771/0.600 for GPT-5.4 vs.\ Grok-4.3. Top-3 overlap ranges from 2/3 to 3/3 and Top-5 overlap from 4/5 to 5/5.
A 10,000-replicate task-paired bootstrap further preserves the winner in 98.4\%, 93.3\%, and 99.8\% of resamples under Gemini-2.5-Pro, GPT-5.4, and Grok-4.3, respectively. Winner-rank 95\% bootstrap intervals are $[1,1]$, $[1,2]$, and $[1,1]$, with mean resampled-ranking Spearman correlations of 0.803, 0.954, and 0.888 against the official ordering restricted to the matched systems. In the 14-task Research stratum, winner preservation is 84.5\%, 97.0\%, and 97.4\%; the six-task Daily stratum is too small for stable stratified estimates. Middle-rank intervals are wide.

\noindent \textbf{Same-family score differentials.}
For reports from a judge's own model family, we compare that judge's score with the mean score from the other two judges; negative values indicate lower own-family scores. The mean differences are -3.41 points for Gemini, -2.53 for GPT, and +1.61 for Grok. On this matched subset, the Gemini differential does not show an own-family score uplift; the mixed signs do not support a uniform self-preference pattern across judges.

\noindent \textbf{Holistic judge control and component interpretation.}
On the same 120 frozen reports, plain holistic LLM-as-a-judge controls produce a stable scalar ordering. The modular pipeline additionally provides module- and item-level diagnostics for visual fidelity, citation support, and report-image alignment. Separately, the original human study in Table~\ref{tab:human-consistency} reports higher preference agreement for the full evaluator than for its vanilla-prompt baseline; that baseline and protocol differ from the frozen-report holistic control.
\par

\noindent \textbf{Aggregation-weight sensitivity.}
We perturb the official top-level weights around $(w_F,w_T,w_M)=(0.2,0.5,0.3)$ using six local settings, each transferring 0.10 between one component pair while preserving the published internal TRACE coefficient $\lambda_{\textsc{Vef.}}=0.4$.
The six $(w_F,w_T,w_M)$ settings are $(0.1,0.5,0.4)$, $(0.1,0.6,0.3)$, $(0.2,0.4,0.4)$, $(0.2,0.6,0.2)$, $(0.3,0.4,0.3)$, and $(0.3,0.5,0.2)$.

The component summaries and displayed Overall scores come from separate runs: the former were obtained in a later independently restarted run. Each system is anchored at its historical Overall and receives only a counterfactual reweighting delta from those later components. One system, Gemini 2.5 Pro, uses partly estimated components constrained by displayed submetrics and Overall; the retained exclusion check gives the same qualitative conclusion. This is an aggregate-level reweighting analysis, not a full-pipeline rerun.
\par

Across these six local perturbations, the winner and full Top-3 set are preserved; mean Spearman with the official 28-system leaderboard is 0.978 (minimum 0.964), and mean Top-5 overlap is 4.5/5.
\par

Alternative aggregation on the available components is subject to the same provenance limitation. Geometric, harmonic, and min-gated aggregation preserve the winner and Top-3 set (Spearman 0.938, 0.871, and 0.755). Equal weighting has Spearman 0.967 and swaps only the top two systems, which are separated by 1.18 points in the official table. Less compensatory aggregation gives weak modules greater influence; the lower min-gated correlation illustrates the sensitivity of ordering to this aggregation choice.

\noindent \textbf{FLAE internal-weight sensitivity and component complementarity.}
We perturb each task-adaptive FLAE dimension weight independently by up to $\pm30\%$, renormalize the weights per task, and recompute system rankings over 2,000 randomized perturbations. Mean Spearman correlation with the default FLAE ranking is 0.999 and the minimum is 0.995; uniform and global-mean weighting are similarly stable. Across report-level module scores, the largest absolute pairwise correlation is 0.313. Under the official judge, leave-one-component-out rankings (with remaining top-level weights renormalized) have Spearman correlations of 0.943, 0.829, and 0.771 with the full ranking when removing FLAE, TRACE, and MOSAIC, respectively.

\subsubsection{Intra-Judge Consistency: Full Results}
\label{app:judge-consistency}
We assess repeated-scoring consistency separately from cross-judge and aggregation-weight sensitivity. Gemini-3.1-Pro reports cover 18 tasks with three generation runs each; GPT-5.4 reports cover 17 tasks with three runs each, excluding Q115 because generation failed. Each fixed report is judged five times by Gemini-2.5-Pro with default temperature 0.2 (VQA uses 0.0), yielding 270 and 255 ratings, respectively.

\begin{table}[!htb]
\caption{Repeated-judge consistency for two report generators. ICC denotes ICC(3,1), $\alpha$ is Krippendorff's agreement coefficient, and CV is coefficient of variation.}
\label{tab:consistency-full}
\centering
\small
\setlength{\tabcolsep}{4pt}
\renewcommand{\arraystretch}{1.10}
\begin{tabular}{l ccc ccc}
\toprule
& \multicolumn{3}{c}{Gemini-3.1-Pro reports} & \multicolumn{3}{c}{GPT-5.4 reports} \\
\cmidrule(lr){2-4}\cmidrule(lr){5-7}
Module & ICC & $\alpha$ & CV & ICC & $\alpha$ & CV \\
\midrule
\textsc{FLAE}   & 0.295 & 0.283 & 0.161 & 0.153 & 0.148 & 0.180 \\
\textsc{TRACE} (w/o \textsc{Vef.})  & 0.884 & 0.883 & 0.210 & 0.702 & 0.699 & 0.213 \\
\textsc{Vef.}   & 0.273 & 0.271 & 0.558 & 0.506 & 0.494 & 0.462 \\
\textsc{MOSAIC} & 0.970 & 0.970 & 0.081 & 0.983 & 0.983 & 0.097 \\
\midrule
Overall MMDR    & 0.709 & 0.705 & 0.160 & 0.614 & 0.602 & 0.158 \\
\bottomrule
\end{tabular}

\end{table}

\noindent \textbf{Decision-level repeatability and interpretation.}
On the 54 Gemini report instances with five repeated ratings each, \textsc{Vef.}'s binary PASS/FAIL judgments have 73.3\% mean pairwise agreement. We report this alongside ICC because the thresholded endpoint differs from a continuous score. With each report held fixed, these repeated ratings measure judge variability. Overall agreement is higher than FLAE and VEF agreement. These report-level repeatability results are distinct from system scores averaged across tasks; the task-paired bootstrap above separately assesses system-ranking stability.

\begin{figure*}[!t]
  \centering
  \includegraphics[trim=5bp 70bp 70bp 40bp,clip,width=\textwidth,height=0.70\textheight,keepaspectratio]{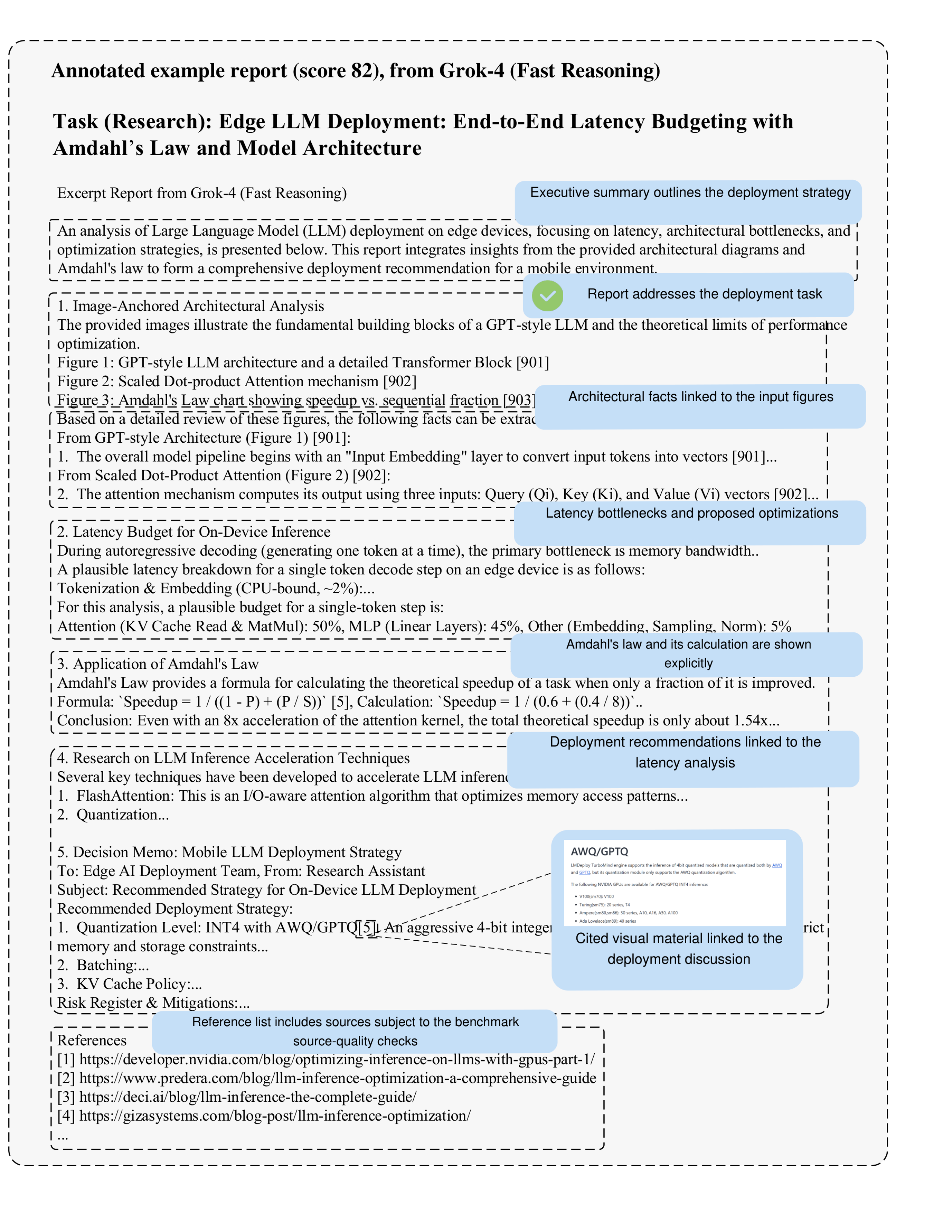}
  \caption{Example scored report generated by Grok-4 (Fast Reasoning).}
  \label{fig:good-report-1}
\end{figure*}

\begin{figure*}[!t]
  \centering
  \includegraphics[trim=35bp 15bp 5bp 30bp,clip,width=0.94\textwidth,height=0.77\textheight,keepaspectratio]{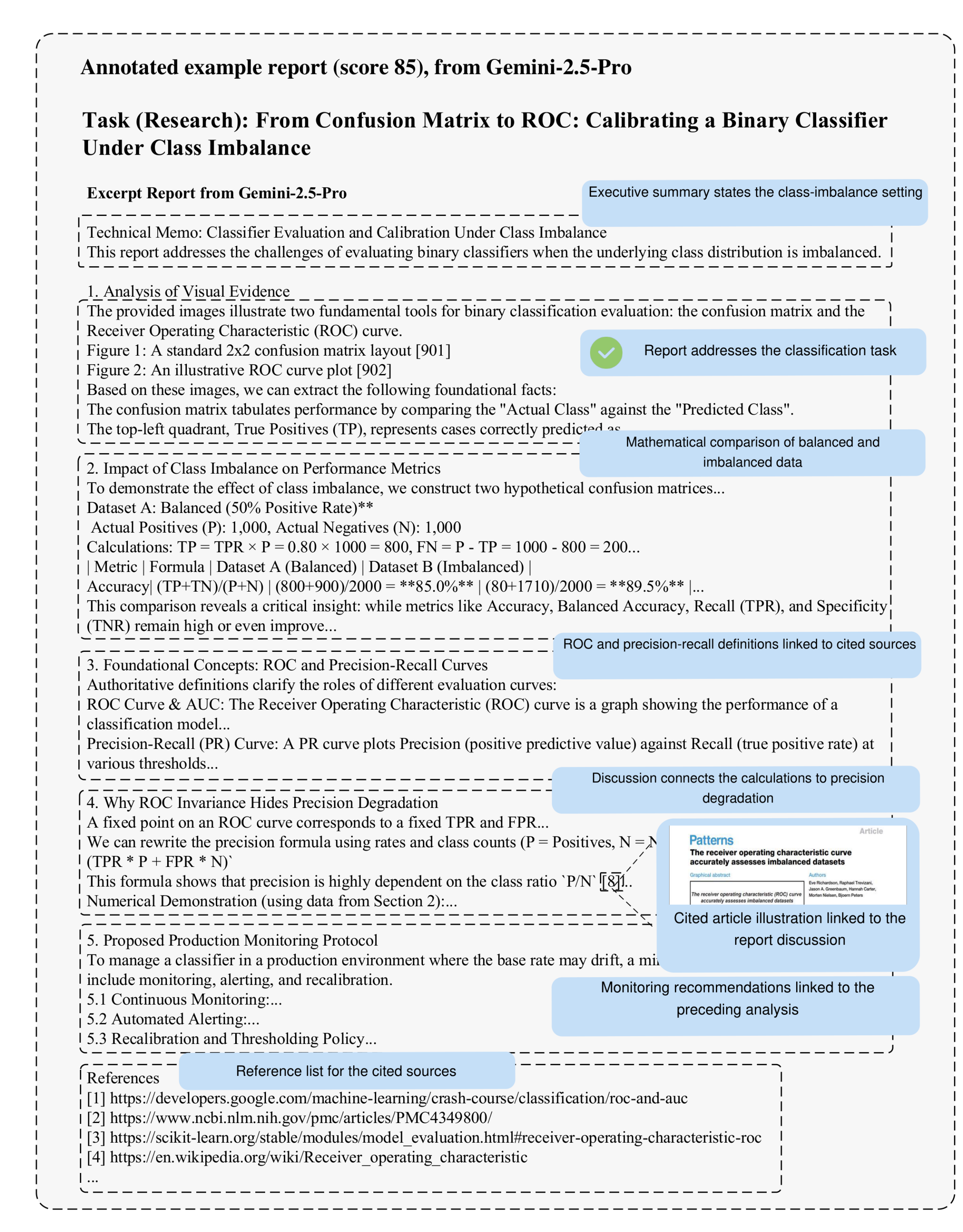}
  \caption{Example scored report generated by Gemini-2.5-Pro.}
  \label{fig:bad-report}
\end{figure*}

Figures~\ref{fig:good-report-1} and~\ref{fig:bad-report} provide annotated report examples from Grok-4 (Fast Reasoning) and Gemini-2.5-Pro; Figure~\ref{fig:vef-failure-example} adds a retained VEF FAIL case illustrating image-reference misalignment.

\subsection{Extended Limitations and Mitigations}
\label{app:limitations}
\textbf{Evaluation scope and coverage.} MMDR-Bench evaluates support relationships between final reports, images, and cited sources. Independent factual verification and causal attribution to retrieval or reasoning stages require additional evidence and protocols. Coverage is defined over extracted key textual claims, and source diversity describes the cited-source distribution. The benchmark is predominantly English and Chinese. Separate module diagnostics, supplementary blinded audits, and documented evaluator-version and coverage conventions support interpretation of these scores (Appendices~\ref{app:flae-formulas}, \ref{app:reliability}, and~\ref{app:human-protocol}).

\textbf{Dynamic services and public tasks.} Retained reports and evidence enable frozen-artifact rescoring. Regeneration also depends on live web access and platform-managed model backends. Public task exposure introduces a risk of memorized task framings; held-out tasks, expanded language coverage, and release versioning are planned maintenance directions.

\textbf{Judge variability.} Appendix~\ref{app:judge-consistency} reports distinct repeatability levels for MOSAIC, FLAE, and VEF. VEF's hard threshold is sensitive to borderline judgments. System-ranking sensitivity and individual-report repeatability characterize complementary evaluation levels; repeated judging can quantify uncertainty at additional evaluation cost.

\end{document}